\documentclass[journal]{IEEEtran}
\usepackage{epsfig}
\usepackage{bm}
\usepackage{hyperref}
\usepackage{amsmath,amssymb}
\usepackage{stmaryrd}
\usepackage{overpic}
\usepackage[linesnumbered,ruled]{algorithm2e}
\usepackage{threeparttable}
\usepackage{booktabs}
\usepackage{multirow}
\usepackage{makecell}
\usepackage{adjustbox}
\usepackage{balance}
\usepackage{graphicx}
\usepackage{cite}
\usepackage[numbers]{natbib}
\begin{document}
\title{Fast Learnings of Coupled Nonnegative Tensor Decomposition Using Optimal Gradient and Low-rank Approximation}

\author{Xiulin~Wang,~\IEEEmembership{}
        Jing~Liu~\IEEEmembership{}
        and~Fengyu~Cong~\IEEEmembership{}
        
\thanks{Corresponding authors: Jing Liu, email: liujing@dmu.edu.cn; Fengyu Cong, email: cong@dlut.edu.cn}% <-this % stops a space
\thanks{X. Wang and J. Liu are with the Stem Cell Clinical Research Center, The First Affiliated Hospital of Dalian Medical University, Dalian 116011, China and also Dalian Innovation Institute of Stem Cell and Precision Medicine, Dalian 116085, China. X. Wang and F. Cong are with the Faculty of Information Technology, University of Jyv{\"a}skyl{\"a}, Jyv{\"a}skyla{\"a} 40100, Finland. F. Cong is also with School of Biomedical Engineering, Faculty of Medicine, Dalian University of Technology, Dalian, China and Key Laboratory of Social Computing and Cognitive Intelligence (Dalian University of Technology), Ministry of Education,China. Xiulin Wang: xiulin.wang@foxmail.com}
\thanks{Manuscript received Month Day, Year; revised Month Day, Year.}}

\markboth{JOURNAL OF LATEX CLASS FILES,~Vol.~*, No.~*, Month~Year}%
{Wang \MakeLowercase{\textit{et al.}}: Fast Learnings of Coupled Nonnegative Tensor Decomposition Using Optimal Gradient and Low-rank Approximation}



\maketitle

\begin{abstract}
Tensor decomposition is a fundamental technique widely applied in signal processing, machine learning, and various other fields. However, traditional tensor decomposition methods encounter limitations when jointly analyzing multi-block tensors, as they often struggle to effectively explore shared information among tensors. In this study, we first introduce a novel coupled nonnegative CANDECOMP/PARAFAC decomposition algorithm optimized by the alternating proximal gradient method (CoNCPD-APG). This algorithm is specially designed to address the challenges of jointly decomposing different tensors that are partially or fully linked, while simultaneously extracting common components, individual components and, core tensors. Recognizing the computational challenges inherent in optimizing nonnegative constraints over high-dimensional tensor data, we further propose the lraCoNCPD-APG algorithm. By integrating low-rank approximation with the proposed CoNCPD-APG method, the proposed algorithm can significantly decrease the computational burden without compromising decomposition quality, particularly for multi-block large-scale tensors. 
Simulation experiments conducted on synthetic data, real-world face image data, and two kinds of electroencephalography (EEG) data demonstrate the practicality and superiority of the proposed algorithms for coupled nonnegative tensor decomposition problems. Our results underscore the efficacy of our methods in uncovering meaningful patterns and structures from complex multi-block tensor data, thereby offering valuable insights for future applications.

\end{abstract}

\begin{IEEEkeywords}
Alternating proximal gradient, CANDECOMP/PARAFAC, coupled tensor decomposition, low-rank approximation, nonnegative tensor decomposition
\end{IEEEkeywords}

% \fi
%
% For peer review papers, this IEEEtran command inserts a page break and
\IEEEpeerreviewmaketitle

\section{Introduction}
-

\IEEEPARstart{D}{ecomposing} a tensor into a minimal number of rank-1 tensors is known as CANDECOMP/PARAFAC (or Canonical Polyadic, \cite{hitchcock1927expression,harshman1970foundations,carroll1970analysis}) decomposition (CPD). Nonnegative CPD (NCPD) imposes nonnegative constraints on its latent factors, providing a part-based representation of a tensor that can extract more meaningful and convincing information \cite{cichocki2009nonnegative}. For instance, electroencephalography (EEG) data, characterized by spatial, temporal, and subject dimensions, can be represented as a third-order tensor, with underlying features extractable through CPD \cite{cong2015tensor}. When considering the time-frequency representation, non-negativity will be naturally brought into the EEG data, making NCPD a necessary method\cite{cong2013multi}. CPD and NCPD methods have garnered significant attention in both theory and application, primarily focusing on the analysis of tensors from a single dataset \cite{kolda2009tensor,cichocki2015tensor,sidiropoulos2017tensor}. When it comes to emerging multi-block tensors, such as multi-subject or multi-modal biomedical data that require joint analysis, traditional tensor decomposition methods become quite challenging in effectively identifying the interconnections among different tensors \cite{zhou2016linked, wang2020group}. For example, traditional methods are inadequate for resolving the internal sharing within EEG data from multiple subjects, or integrating the complementary spatiotemporal characteristics in EEG-fMRI data \cite{wang2020group,jonmohamadi2019extraction}. 

Coupled tensor decomposition, an extension of tensor decomposition to multi-block tensors, allows for the simultaneous decomposition of two or more tensors with an enforced shared structure \cite{zhou2016linked,gong2018double,sorensen2015coupled}. Notably, the uniqueness condition for coupled tensor decomposition is more relaxed compared to that of single tensor decomposition \cite{sorensen2015coupled,hunyadi2017tensor}. This method can leverage various constraints, such as sparsity, smoothness, and nonnegativity, to obtain more unique solutions and interpretable components \cite{xue2019coupled, cichocki2013tensor}. By preserving the multi-way structure of tensor data, Coupled tensor decomposition can achieve higher decomposition accuracy and robustness, revealing interconnections among tensor data that would inevitably be lost in two-way matrix counterparts \cite{cong2015tensor,xue2019coupled,wang2020group,calhoun2009review,gong2015generalized}, while coupled tensor methods can circumvent the independent constraint \cite{hunyadi2017tensor,morup2011applications}. Moreover, coupled tensor decomposition of multi-block tensors can achieve the simultaneous extraction of common components shared by all tensors and individual components specific to each tensor \cite{yokota2012linked,wang2019generalization,zhou2015group}.

To date, increasing recognition of joint tensor analysis has spurred the exploration of coupled tensor decomposition in various applications. In certain cases, coupled matrix and tensor factorization (CMTF \cite{acar2011all}) and its derivatives \cite{acar2017acmtf,jonmohamadi2019extraction, chatzichristos2018fusion,schenker2023parafac2}, such as advanced CMTF \cite{acar2017acmtf} and coupled tensor-tensor decomposition (CTTD \cite{jonmohamadi2019extraction, chatzichristos2018fusion}), have demonstrated their superiority over ICA-based two-way methods in fusing EEG and fMRI data. The algebraic double-coupled CPD (DC-CPD) algorithm, which uses second-order statistics in joint blind source separation problems, exhibits greater uniqueness and accuracy than the standard CPD \cite{gong2018double}. Linked CPD models, optimized by hierarchical alternating least squares (HALS), fast HALS, and alternating direction method of multipliers (ADMM), have also achieved good performance in classification, image processing, and biomedical signal processing \cite{wang2020group,yokota2012linked,wang2021shared,zdunek2019linked}. Additionally, the common and individual feature extraction (CIFE, \cite{zhou2015group}) framework for multi-block data facilitates the separate extraction of common and individual components by incorporating dimension reduction and blind source separation (BSS) methods, and has been successfully applied to classification, clustering and linked BSS problems \cite{zhou2016linked}. Coupled tensor decomposition also finds applications in data fusion of low spatial resolution hyperspectral (LRHS) and high spatial resolution multispectral (HRMS) images \cite{kanatsoulis2018hyperspectral,li2018fusing,xu2024coupled}, array signal processing \cite{sorensen2013coupled,sorensen2016multidimensional1,sorensen2016multidimensional2}, linked prediction\cite{ermics2015link} and metabolic physiology \cite{acar2015data}.

Nevertheless, existing coupled tensor decomposition methods often face challenges such as slow convergence speed and low optimization accuracy, primarily arising from the nonnegative constraint and the high-dimensional nature of tensor data \cite{zhou2012fast,zhang2016fast}. Therefore, aiming to address these challenges and achieve effective and efficient joint tensor analysis with coupled information, we propose two advanced coupled NCPD methods: the coupled nonnegative CANDECOMP/PARAFAC decomposition algorithm based on alternating proximal gradient (CoNCPD-APG) and its fast implementation incorporating low-rank approximation (lraCoNCPD-APG). Specifically, our contributions in this study are outlined as follows:
\begin{enumerate}
    \item Using the optimal gradient method, we propose an effective CoNCPD-APG algorithm for joint analysis of partially or fully linked multi-block tensors, which enables simultaneous decomposition of tensors with excellent decomposition accuracy.
    \item By introducing low-rank approximation, we further propose an efficient larCoNCPD-APG algorithm, which can greatly reduce time consumption without losing decomposition accuracy.
    \item We thoroughly analyze the types of couplings, algorithmic error bound and computational complexity, while also providing an insight into calculating the number of coupled tensor components.
    \item The experiments designed on synthetic data, real-world face image and two types of EEG data prove the practicability and superiority of the proposed algorithms. 
\end{enumerate}
The rest of this paper is organized as follows. Section~\ref{sec:prelirework} introduces some basic preliminaries and related work. In Section~\ref{se:section3}, we present the proposed algorithms as well as some theoretical analyses. Experiments on synthetic and real-world data are performed in Section~\ref{sec:exp} to verify the performance of the proposed algorithms.
The last section concludes this paper.
\section{Preliminaries and related work}
\label{sec:prelirework}
\subsection{Notations and tensor operations}
Tensor, also known as multi-way array, is the high-order generalization of vector and matrix. The order of a tensor is the number of dimensions, ways or modes of itself. In this study, tensors are represented by calligraphic boldface uppercase letters, matrices by boldface uppercase letters, vectors by boldface uppercase letters, and scalars by lowercase letters. Table~\ref{mathnotations} gives a summary of basic notations and mathematical operations throughout this study, and please refer to \cite{kolda2009tensor} for a more detailed description of them.
\begin{table}[h]
\renewcommand\arraystretch{1.2}
\caption{Basic notations and mathematical operations}
\label{mathnotations}
\centering
\begin{threeparttable}
\begin{tabular}{ll}
    \hline  
		{\bf  Symbol}& {\bf Defination}\\  
    \hline			
		$\mathbb{R}$, $\mathbb{R}_+$&real number, nonnegative real number\\		
		$\circ$, $\odot$& outer product, Khatri-Rao product\\		
		$\circledast$, $\oslash$ &element-wise (Hadamard) product, division\\
		$(\cdot)^T$, $\textrm{vec}(\cdot)$&transpose, vectorization operator\\
		$\left\llbracket\cdot\right\rrbracket$, $\left\langle\cdot,\cdot\right\rangle$, $\left\|\cdot\right\|_{F}$&Kruskal operator, inner product and Frobenius norm\\
		$m$, $\bm m$, $\bm M$, $\bm{\mathcal M}$ & scalar, vector, matrix and tensor\\	
		$\bm{\mathcal M}_{(n)}$& mode-$n$ matricization of tensor $\bm{\mathcal M}$\\	
		$\textrm{vec}(\bm{\mathcal M})$& vectorization of tensor $\bm{\mathcal M}$\\
		$\textrm{ddiag}(\bm{\mathcal M})$&vectorization of super-diagonal elements of tensor $\bm{\mathcal M}$\\
		$\textrm{ddiag}(\bm m)$&tensorization with $\bm m$ on the super-diagonal elements\\		
		$\bm U_{i:j,:}$&row-wise submatrix of $\bm U$, $i$th to $j$th row\\
		$\bm U_{:,i:j}$&column-wise submatrix of $\bm U$, $i$th to $j$th column\\
		$\bm U^{(n)}$&the $n$-th factor matrix\\
		$\;\bm U^{\odot}$&$\bm{U}^{(N)}\odot\bm{U}^{(N-1)}\odot\cdots\odot\bm{U}^{(2)}\cdots\odot\bm{U}^{(1)}$ \\
		$\;\bm U^{\odot_{-n}}$&$\bm{U}^{(N)}\odot\cdots\odot\bm{U}^{(n+1)}\odot\bm{U}^{(n-1)}\cdots\odot\bm{U}^{(1)}$ \\	
            $\;\bm U^{\circledast}$&$\bm{U}^{(N)}\circledast\bm{U}^{(N-1)}\circledast\cdots\circledast\bm{U}^{(2)}\cdots\circledast\bm{U}^{(1)}$ \\
		$\;\bm U^{\circledast_{-n}}$&$\bm{U}^{(N)}\circledast\cdots\circledast\bm{U}^{(n+1)}\circledast\bm{U}^{(n-1)}\cdots\circledast\bm{U}^{(1)}$ \\	
    \hline   
\end{tabular}
\end{threeparttable}
\end{table}

\subsection{Optimal gradient method}
Accelerated/Alternating proximal gradient (APG), an accelerated version of proximal gradient (PG, \cite{parikh2014proximal}), was originally proposed by Nesterov for smooth optimization with achieving the convergence rate of $\mathcal{O}(\frac{1}{K^2})$, where $K$ is the number of iterations \cite{nesterov1983method,beck2009fast}. For a minimization problem: $\textrm{min}\{f(x), x\in\mathbb{R}^n\}$, assuming that $f(x):\mathbb{R}^n\rightarrow\mathbb{R}$ is a convex function with Lipschitz continuous gradient $f'$, there will hold that
\begin{equation}
\left\|f'(x_i)-f'(x_j)\right\|\leq L \left\|x_i-x_j\right\|, \forall x_i,x_j\in\mathbb{R}^n,
\end{equation} 
where $L>0$ is the Lipschitz constant. To obtain an optimal point $\ddot{x}$, two sequences are updated successively in each iteration round (assume at the $k$th iteration) in APG method \cite{nesterov1983method, guan2012nenmf} as follows:
\begin{equation}
\label{eq:proxifunc}
\begin{aligned}
x_k=\underset{x}{\textrm{argmin}}\bigg\{\phi(x,x_{k-1})=f(x_{k-1})\bigg.\\\left.+\left\langle x-x_{k-1},f'(x_{k-1})\right\rangle+\frac{L}{2}\left\|x-x_{k-1}\right\|\right\},
\end{aligned}
\end{equation}
and
\begin{equation}
\label{eq:extrapp}
x_{k+1}=x_{k}+\frac{\alpha_k-1}{\alpha_{k+1}}(x_k-x_{k-1})
\end{equation}
with
\begin{equation}
\alpha_{0}=1,\;\alpha_{k+1}=\frac{1+\sqrt{4\alpha_k^2+1}}{2}
\end{equation}
where $\phi(x,x_{k-1})$ denotes the proximal regularized function of $f(x)$ at $x_{k-1}$ and (\ref{eq:extrapp}) denotes an extrapolated point by combining the points of current and previous iterations. Using the Lagrange multiplier method, from (\ref{eq:proxifunc}), we have 
\begin{equation}
x_{k}\leftarrow \mathcal{P}(x_{k-1}-\frac{1}{L}f'(x_{k-1}))
\end{equation}
where $\mathcal{P}(\cdot)$ denotes a shrinkage operator predefined by the user. Since each subproblem under block coordinate descent (BCD) framework is a convex function with Lipchitz continuous gradient, APG and its variants have proven to be very efficient for nonnegative matrix/tensor factorization issues and outperform many other competitors \cite{zhang2016fast,guan2012nenmf,xu2013block,xu2015alternating}. In the sequel, we introduce the APG method to solve the coupled nonnegative tensor decomposition problems.
\subsection{Coupled NCPD model}
In this study, we mainly consider the coupled NCPD model, given a set of $N$th-order nonnegative tensors $\bm{\mathcal{M}}^{(s)}\in\mathbb{R}^{I_1\times I_2\times \cdots I_N }_+$, $s=1,2,\cdots,S$, the model can be expressed as:
\begin{equation}
\label{eq:tensormodel}
\begin{split}
\bm{\mathcal{M}}^{(s)}\! \approx\! \bm{\mathcal{\hat{M}}}^{(s)}&= \sum_{r=1}^{R^{(s)}}\lambda^{(s)} _{r}\bm{\mathit{u}}_{r}^{(1,s)}\circ\bm{\mathit{u}}_{r}^{(2,s)}\circ\cdots\circ \bm{\mathit{u}}_{r}^{(N,s)} \\&= \left\llbracket \bm{\mathcal{D}}^{(s)};\bm{\mathit{U}}^{(1,s)},\bm{\mathit{U}}^{(2,s)},\cdots,\bm{\mathit{U}}^{(N,s)} \right\rrbracket,
\end{split}
\end{equation}
where $\bm{\mathcal{\hat{M}}}^{(s)}\in \mathbb{R} ^{I_1\times I_2\times \cdots I_N }_+$ denotes the estimation of $\bm{\mathcal{M}}^{(s)}$. $\bm{u}_{r}^{(n,s)}\in\mathbb{R}_+^{I_n}$ denotes the $r$th column of mode-$n$ factor matrix of the $s$th tensor and $\bm{U}^{(n,s)}=\left[ \bm{u}_{1}^{(n,s)},\bm{u}_{2}^{(n,s)},\cdots,\bm{u}_{R}^{(n,s)}\right]\in\mathbb{R}_+^{I_n\times {R^{(s)}}}$. $\bm{\mathcal{D}}^{(s)}\in\mathbb{R}_+^{{R^{(s)}}\times\cdots\times {R^{(s)}}}$ represents the $s$th core tensor with non-zero entries $\lambda_{r}^{(s)}$ only on its super-diagonal elements. $\bm{\mathcal{\hat{M}}}_r^{(s)}=\lambda^{(s)} _{r}\bm{\mathit{u}}_{r}^{(1,s)}\circ\bm{\mathit{u}}_{r}^{(2,s)}\circ\cdots\circ \bm{\mathit{u}}_{r}^{(N,s)}$ is termed as a rank-1 tensor generated by the outer product of $\bm{u}_{r}^{(n,s)}$, $n=1,2,\cdots,N$, and $\lambda_r^{(s)}$ is used to represent the scaling of rank-1 tensor. The decomposition of each tensor $\bm{\mathcal{M}}^{(s)}$ can be regarded as decomposing a high-order tensor into a minimal number of rank-1 tensors, and the minimum number $R^{(s)}$ is termed as the rank of the tensor or the number of components.

In coupled NCPD model, we assume that each factor matrix includes two parts and satisfies $\bm{U}^{(n,s)}=\left [\bm{U}_{C}^{(n,s)}\;\bm{U}_{I}^{(n,s)}\right ]$. $\bm{U}_{C}^{(n,s)}\in \mathbb{R}_+^{I_{n}\times L_{n}}$, $0 \leq  L_{n} \leq \textrm{min}(R^{(s)})$ represents the common information shared by all tensors as $\bm{U}_{C}^{(n,1)}=\cdots=\bm{U}_{C}^{(n,S)}=\bm{U}_{C}^{(n)}$, and $\bm{U}_{I}^{(n,s)}\in \mathbb{R}_+^{I_{n}\times ({R^{(s)}}-L_{n})}$ denotes the individual part corresponding to individual tensor. $L_n$ represents the number of coupled components between tensors in the $n$th mode. 
\section{Proposed Algorithm}
\label{se:section3}
This section illustrates how to use APG method or combine APG method and low-rank approximation to solve the coupled NCPD problem. In addition, we give some discussions on the properties of the proposed algorithms as well as some implementation remarks.
\subsection{Coupled NCPD using APG} 
For the coupled NCPD model, the optimization criterion of Euclidean divergence minimization is adopted to
minimize the error between the original and estimated tensors. Therefore, given a set of nonnegative tensors $\bm{\mathcal{M}}^{(s)}$, $s=1,2,\dots,S$, the objective function of coupled NCPD model can be presented as follows:
\begin{equation}
\label{eq:costfunc}
\underset{\bm{\mathcal{D}}^{(s)},\bm{U}^{(n,s)}}{\textrm{min}\; F}\frac{1}{2}\sum_{s=1}^{S}\left \| \bm{\mathcal{M}}^{(s)} -\left\llbracket \bm{\mathcal{D}}^{(s)};\bm{U}^{(1,s)},\cdots,\bm{U}^{(N,s)} \right\rrbracket \right \|_{F}^{2}
\end{equation}
\[
\textrm{s.t.,}\;\bm{\mathcal{D}}^{(s)}\in\mathbb{R}_+^{{R^{(s)}}\times\cdots\times {R^{(s)}}},\;\bm{U}^{(n,s)}\in\mathbb{R}_+^{I_n\times R^{(s)}}
\]
where $\bm{U}^{(n,s)}=\left [\bm{U}_{C}^{(n,s)}\;\bm{U}_{I}^{(n,s)}\right ]$ and $\bm{U}_{C}^{(n,1)}\!=\!\cdots\!=\!\bm{U}_{C}^{(n,S)}\!=\!\bm{U}_{C}^{(n)}$.
According to BCD framework, the coupled NCPD problem can be converted into several subproblems by optimizing $\bm{\mathcal D}^{(s)}$ and $\bm{U}^{(n,s)}$ alternatively in each iteration. Each subproblem can be regarded as a minimization problem of a continuously differentiable function, which can be solved efficiently by APG methed \cite{guan2012nenmf,xu2013block,xu2015alternating}. Next we provide a solution for coupled NCPD problem based on APG method. 

First, regarding the solution of core tensor $\bm{\mathcal{D}}^{(s)}$, we adopt (\ref{eq:eqsolD}) (see it at the top of next page). By keeping all the other variables and using the Lagrange multiplier method,  $\bm{\mathcal{D}}^{(s)}$ can be updated in a closed form as follows:
\begin{figure*}[!t]
\begin{equation}
\label{eq:eqsolD}
\bm{\mathcal D}^{(s)}=\underset{\bm{\mathcal D}^{(s)}\geq0}{\textrm{argmin}}\left [ F(\hat{\bm{\mathcal D}}^{(s)})+\left \langle \hat{\bm {\mathcal G}}^{(s)}, \bm{\mathcal D}^{(s)}-\hat{\bm{\mathcal D}}^{(s)}\right \rangle + \frac{L_d^{(s)}}{2} \left \| \bm{\mathcal D}^{(s)}-\hat{\bm{\mathcal D}}^{(s)} \right \| ^{2}_F\right]
\end{equation}
\hrulefill
\end{figure*}
\begin{equation}
\label{eq:solD}
\bm{\mathcal D}^{(s)}=\textrm{max}\left(0,\;\hat{\bm{\mathcal D}}^{(s)}-\frac{\hat{\bm {\mathcal G}}^{(s)}}{L_d^{(s)}}\right)
\end{equation}
where $\bm{\hat{\mathcal D}}^{(s)}$ denotes an extrapolated point and $L_d^{(s)}$ denotes the Lipschitz constant of $F'(\bm{\mathcal D}^{(s)})$. $\hat{\bm {\mathcal G}}^{(s)}$ is the block-partial gradient of (\ref{eq:costfunc}) at $\hat{\bm{\mathcal D}}^{(s)}$, which can be calculated as:
\begin{equation}
\begin{split}
\label{eq:solapgG}
\hat{\bm {\mathcal G}}^{(s)}=\textrm{ddiag}\left[\left(\bm{U}^{{(s)}^T}\bm U^{(s)}\right)^\circledast\textrm{ddiag}\left(\hat {\bm {\mathcal D}}^{(s)}\right)\right.\\\left.-\left(\bm{U}^{(s)^\odot}\right)^T\textrm{vec}\left(\bm{\mathcal{M}}^{(s)}\right)\right]
\end{split}
\end{equation}
where $\textrm{vec}(\bm{\mathcal M}^{(s)})$ denotes the vectorization of tensor $\bm{\mathcal M}^{(s)}$. $\textrm{ddiag}(\bm{\mathcal{D}}^{(s)})$ denotes a vector vectorized from the super-diagonal elements of $\bm{\mathcal{D}}^{(s)}$, and the outer-loop notation $\textrm{ddiag}(\cdot)$ means the tensorization from a vector to a super-diagonal tensor, which is the reverse operation of the inner-loop $\textrm{ddiag}(\cdot)$.

Second, for the solution of factor matrix $\bm{U}^{(n,s)}$ (without coupled information), we adopt the updating method as (\ref{eq:eqsolUns}) (see it at the top of next page), which can be written in the closed form as follows:
\begin{figure*}[!t]
\begin{equation}
\label{eq:eqsolUns}
\bm{U}^{(n,s)}=\underset{\bm{U}^{(n,s)}\geq0}{\textrm{argmin}}\sum_{s=1}^{S}\left [F(\hat{\bm{U}}^{(n,s)})+\left \langle \hat{\bm G}^{(n,s)}, \bm{U}^{(n,s)}-\hat{\bm{U}}^{(n,s)}\right \rangle + \frac{L_{u}^{(n,s)}}{2} \left \| \bm{U}^{(n,s)}-\hat{\bm{U}}^{(n,s)} \right \| ^{2}_F\right]
\end{equation}
\hrulefill
\end{figure*}
\begin{equation}
\label{eq:solUns}
\bm U^{(n,s)}=\textrm{max}\left(0,\;\hat{\bm U}^{(n,s)}-\frac{\hat{\bm G}^{(n,s)}}{L_u^{(n,s)}}\right)
\end{equation}
where $\bm{\hat U}^{(n,s)}$ denotes an extrapolated point of $\bm{U}^{(n,s)}$, $L_{u}^{(n,s)}$ denotes a Lipschitz constant of block-partial gradient of (\ref{eq:costfunc}) at $\bm{U}^{(n,s)}$. The block-partial gradient $\hat{\bm G}^{(n,s)}$ of (\ref{eq:costfunc}) at $\hat{\bm U}^{(n,s)}$ is expressed as:
\begin{equation}
\begin{split}
\label{eq:solapgU}
\hat{\bm G}^{(n,s)}&=\hat{\bm U}^{(n,s)}\bm D^{(s)}\left(\bm{U}^{{(s)}^T}\bm U^{(s)}\right)^{\circledast_{-n}}\bm D^{{(s)}}\\&-\bm{\mathcal M}_{(n)}^{(s)}\bm{U}^{{(s)}^{\odot_{-n}}}\bm D^{{(s)}}
\end{split}
\end{equation}
where $\bm{\mathcal{M}}_{(n)}^{(s)}$ denotes the mode-$n$ matricization of $\bm{\mathcal{M}}^{(s)}$. $\bm D^{(s)}$ is a diagonal matrix and its diagonal elements correspond to the super-diagonal elements of core tensor $\bm {\mathcal{D}}^{(s)}$. 

Third, for the factor matrix which includes $\bm{U}_{C}^{(n)}$ and $\bm{U}_{I}^{(n,s)}$, we need to calculate their solutions separately. Since $\bm{U}_C^{(n)}$ is shared by all tensors as $\bm{U}_{C}^{(n,1)}\!=\!\cdots\!=\!\bm{U}_{C}^{(n,S)}\!=\!\bm{U}_{C}^{(n)}$, we should combine the information of all tensors to calculate the solution of $\bm{U}_C^{(n)}$. The solution of individual part $\bm{U}_{I}^{(n,s)}$ only needs to consider the corresponding $s$th-set tensor. Therefore, we have
\begin{equation}
\label{eq:solUnC}
\bm U^{(n)}_{C}=\textrm{max}\left(0,\;\hat{\bm U}^{(n)}_{C}-\frac{\sum_{s=1}^{S}\hat{\bm G}_{C}^{(n,s)}}{\sum_{s=1}^{S}L_u^{(n,s)}}\right),
\end{equation}
and
\begin{equation}
\label{eq:solUnsI}
\bm U^{(n,s)}_{I}=\textrm{max}\left(0,\;\hat{\bm U}^{(n,s)}_{I}-\frac{\hat{\bm G}_{I}^{(n,s)}}{L_u^{(n,s)}}\right)
\end{equation}
where $\hat{\bm G}_{C}^{(n,s)}$ and $\hat{\bm G}_{I}^{(n,s)}$ denote the block-partial gradients of (\ref{eq:costfunc}) at $\bm{\hat U}_C^{(n,s)}$ and $\bm{\hat U}_I^{(n,s)}$, respectively. $\bm{\hat U}_C^{(n,s)}$ and $\bm{\hat U}_I^{(n,s)}$ denote 
the extrapolated points of $\bm{U}_C^{(n,s)}$ and $\bm{U}_I^{(n,s)}$. Moreover, $\bm{\hat U}^{(n,s)}=\left[\bm{\hat U}_C^{(n,s)}\;\bm{\hat U}_I^{(n,s)}\right]$ and $\bm{\hat G}^{(n,s)}=\left[\bm{\hat G}_C^{(n,s)}\;\bm{\hat G}_I^{(n,s)}\right]$. 

Consider updating $\bm{\mathcal D}^{(s)}$ and $\bm{U}^{(n,s)}$ at the $k$th iteration. The extrapolated points $\bm{\hat{\mathcal D}}_{k-1}^{(s)}$ and $\bm{\hat U}_{k-1}^{(n,s)}$ are defined as
\begin{equation}
\label{eq:eqextraPG}
\hat{\bm {\mathcal D}}^{(s)}_{k-1}={\bm {\mathcal D}}^{(s)}_{k-1}+w^{(s)}_{d,k-1}\left({\bm{\mathcal D}}^{(s)}_{k-1}-{\bm{\mathcal {D}}}^{(s)}_{k-2}\right),
\end{equation}
and 
\begin{equation}
\label{eq:eqextraPU}
\hat{\bm U}^{(n,s)}_{k-1}={\bm U}^{(n,s)}_{k-1}+w^{(n,s)}_{u,k-1}\left({\bm U}^{(n,s)}_{k-1}-{\bm U}^{(n,s)}_{k-2}\right)
\end{equation}
where $w^{(s)}_{d,k-1}$ and $w^{(n,s)}_{u,k-1}$ denote the extrapolation weights. Since APG is not a monotone method, i.e., $F(k)$ may not be smaller than $F(k\!-\!1)$. Therefore, if $F(k)\geq F(k\!-\!1)$ after iteration $k$, an additional re-updating of ${\bm U}^{(n,s)}_{k}$ and ${\bm {\mathcal D}}^{(s)}_{k}$ will be taken via $\hat{\bm{\mathcal D}}_{k-1}^{(s)}=\bm{\mathcal D}_{k-1}^{(s)}$ and	$\hat{\bm{U}}_{k-1}^{(n,s)}=\bm{U}_{k-1}^{(n,s)}$. In each iteration, we perform the optimization with the order $\bm{\mathcal D}^{(1)},\bm{\mathcal D}^{(2)},\cdots,\bm{\mathcal D}^{(S)}$ and $\bm{U}^{(1,1)},\cdots\bm{U}^{(1,S)},\cdots,\bm{U}^{(N,1)},\cdots\bm{U}^{(N,S)}$, which are alternatively updated one after another until convergence. We term the proposed coupled NCPD algorithm based on APG update as CoNCPD-APG and summarize it in Algorithm \ref{lst:CoNTD-APG}. The detailed derivations and relevant parameter settings are given in the Appendix \ref{app:appenA}.

\begin{algorithm}[]
      \caption{CoNCPD-APG algorithm}
      \label{lst:CoNTD-APG}
      \KwIn{$\bm{\mathcal{M}}^{(s)}$, $L_n$ and $R^{(s)}$, $n\!=\!1,\!\cdots\!,N$, $s\!=\!1,\!\cdots\!,S$}          
      Initialization:\\ $\bm {U}^{(n,s)}$,
$\bm{\mathcal D}^{(s)}$, $\bm{\mathcal{M}}_{(n)}^{(s)}$, $n=1,\cdots,N$, $s=1,\cdots,S$\\
      \For{$k=1,2,\cdots,\textrm{MaxIt}$}
      {     
					\For{$s=1,\cdots,S$}{
					Calculate $\hat{\bm{\mathcal{G}}}_{k-1}^{(s)}$ and $\hat{\bm{\mathcal D}}_{k-1}^{(s)}$ via (\ref{eq:solapgG}) and (\ref{eq:eqextraPG})\\
					Update $\bm{\mathcal D}_k^{(s)}$ via (\ref{eq:solD})
						}
             \For{$n=1,2,\cdots,N$} 
					 {
					 \For{$s=1,\cdots,S$}{
					 Calculate $\hat{\bm G}^{(n,s)}_{k-1}$ and $\hat{\bm U}^{(n,s)}_{k-1}$ via (\ref{eq:solapgU}) and (\ref{eq:eqextraPU})\\				
               Update $\bm {U}_k^{(n,s)}$ via (\ref{eq:solUns}), (\ref{eq:solUnC}) and (\ref{eq:solUnsI})
               }
             }
			\If{${F}(k)\geq{F}(k-1)$}
			{
         $\hat{\bm{\mathcal D}}_{k-1}^{(s)}=\bm{\mathcal D}_{k-1}^{(s)}$,	$\hat{\bm{U}}_{k-1}^{(n,s)}=\bm{U}_{k-1}^{(n,s)}$\\			
				Reupdate $\bm{\mathcal D}_k^{(s)}$, $\bm {U}_k^{(n,s)}$ via (\ref{eq:solD}), (\ref{eq:solUns}), (\ref{eq:solUnC}) and (\ref{eq:solUnsI})
			}
			\If{stopping criteron is satisfied}
			{
				\Return{\\$\bm {U}_k^{(n,s)}$, $\bm{\mathcal D}_k^{(s)}$, $n=1,\cdots,N$, $s=1,\cdots,S$}
			}
      }
     \KwOut{$\bm {U}^{(n,s)}$, $\bm{\mathcal D}^{(s)}$, $n=1,\cdots,N$, $s=1,\cdots,S$}
\end{algorithm} 

\subsection{Coupled NCPD using APG and low-rank approximation}
In CoNCPD-APG algorithm, the time-consumption of updating $\bm{\mathcal D}^{(s)}$ and $\bm{U}^{(n,s)}$ is mainly attributed to the multiplication of $(\bm{U}^{(s)^\odot})^T\textrm{vec}(\bm{\mathcal{M}}^{(s)})$ and $\bm{\mathcal M}_{(n)}^{(s)}\bm{U}^{{(s)}^{\odot_{-n}}}$ in (\ref{eq:solapgG}) and (\ref{eq:solapgU}), and it will be increasingly serious especially for the tensors with large dimensionality. Specifically, in each iteration, let $R^{(s)}=R$, the computational complexity of $(\bm{U}^{(s)^\odot})^T\textrm{vec}(\bm{\mathcal{M}}^{(s)})$ reaches $\mathcal{O}(SR\prod_{n}\!I_n)$ and $\bm{\mathcal M}_{(n)}^{(s)}\bm{U}^{{(s)}^{\odot_{-n}}}$ has the complexity of $\mathcal{O}(NSR\prod_{n}I_n)$.
The applications of low-rank approximation in nonnegative matrix/tensor factorization have demonstrated their performance improvement in terms of computational efficiency while maintaining computational accuracy \cite{zhou2012fast,cong2014low,zhang2016fast}. Generally, the unconstrained CPD of a tensor converges in dozens of iterations and is considered faster than its counterpart with nonnegative constraint. Therefore, to reduce the computational complexity of CoNCPD-APG algorithm, we consider introducing the low-rank approximation of $\bm{\mathcal{M}}^{(s)}$ before performing the actual coupled decomposition. Suppose that $\left\llbracket \tilde{\bm{U}}^{(1,s)},\tilde{\bm{U}}^{(2,s)},\cdots,\tilde{\bm{U}}^{(N,s)} \right\rrbracket$
is the rank-$\tilde{R}^{(s)}$ approximation of $\bm{\mathcal{M}}^{(s)}$ obtained by the unconstrained CPD, $\tilde{\bm{U}}^{(n,s)}\in\mathbb{R}^{I_n\times\tilde R^{(s)}}$, $\tilde{R}^{(s)}\!\leq\! R^{(s)}$, thus the cost function in (\ref{eq:costfunc}) can be represented as the optimization problem in (\ref{eq:cosfunclra}) (see it at the top of next page) with fixed $\tilde{\bm{U}}^{(1,s)},\tilde{\bm{U}}^{(2,s)},\cdots,\tilde{\bm{U}}^{(N,s)}$.  In other words, instead of loading the original tensor $\bm{\mathcal{M}}^{(s)}$ directly into the iterations, we first split the tensor into smaller compressed matrices, such as $\tilde{\bm{U}}^{(1,s)},\tilde{\bm{U}}^{(2,s)},\cdots,\tilde{\bm{U}}^{(N,s)}$, and then bring them into the decomposition iterations, which can greatly reduce the time and space complexities of algorithms\cite{zhou2012fast}. Via low-rank approximation, $\textrm{vec}(\bm{\mathcal{M}}^{(s)})$ and $\bm{\mathcal M}_{(n)}^{(s)}$ in (\ref{eq:solapgG}) and (\ref{eq:solapgU}) can be respectively expressed by $\textrm{vec}(\bm{\mathcal{M}}^{(s)})=\tilde{\bm{U}}^{(s)^\odot}\,\textrm{ddiag}\bm{\mathcal{(I)}}$ and $\bm{\mathcal M}_{(n)}^{(s)}=\tilde{\bm{U}}^{(s,n)}\big(\tilde{\bm{U}}^{{(s)}^{\odot_{-n}}}\big)^T$, and $\bm{\mathcal{I}}\in\mathbb{R}^{\tilde R^{(s)}\times\cdots\times\tilde R^{(s)}}$ denotes a core tensor with all super-diagonal elements being 1.
\begin{figure*}[t]
\begin{equation}
\label{eq:cosfunclra}
\underset{\bm{\mathcal{D}}^{(s)},\bm{U}^{(n,s)}}{\textrm{min}\;\mathcal F}\frac{1}{2}\sum_{s=1}^{S}\bigg \| \left\llbracket \tilde{\bm{U}}^{(1,s)},\tilde{\bm{U}}^{(2,s)},\cdots,\tilde{\bm{U}}^{(N,s)} \right\rrbracket -\left\llbracket \bm{\mathcal{D}}^{(s)};\bm{U}^{(1,s)},\bm{U}^{(2,s)},,\cdots,\bm{U}^{(N,s)} \right\rrbracket \bigg \|_{F}^{2}
\end{equation}
\[
\textrm{s.t.}\;\bm{\mathcal{D}}^{(s)}\in\mathbb{R}_+^{I_1\times\cdots\times I_N},\;\bm{U}^{(n,s)}\in\mathbb{R}_+^{I_n\times R^{(s)}},\; \tilde{\bm{U}}^{(n,s)}\in\mathbb{R}^{I_n\times \tilde R^{(s)}},\;\tilde{R}^{(s)}\leq R^{(s)}
\]
\hrulefill
\end{figure*} 
This thereby leads to 
\begin{equation}
\label{eq:eqlrag}
\begin{split}
\left(\bm{U}^{(s)^\odot}\right)^T\textrm{vec}\left(\bm{\mathcal{M}}^{(s)}\right) &= \left(\bm{U}^{(s)^\odot}\right)^T\tilde{\bm{U}}^{(s)^\odot}\,\textrm{ddiag}\bm{\mathcal{(I)}}\\&
=\left(\bm{U}^{{(s)}^T}\tilde{\bm U}^{(s)}\right)^\circledast\textrm{ddiag}\bm{\mathcal{(I)}}
\end{split}
\end{equation}
and
\begin{equation}
\label{eq:eqlrau}
\begin{split}
\bm{\mathcal M}_{(n)}^{(s)}\bm{U}^{{(s)}^{\odot_{-n}}} &= \tilde{\bm{U}}^{(s,n)}\left(\tilde{\bm{U}}^{{(s)}^{\odot_{-n}}}\right)^T\bm{U}^{{(s)}^{\odot_{-n}}}\\&
=\tilde{\bm{U}}^{(s,n)}\left(\tilde{\bm{U}}^{{(s)}^T}\bm U^{(s)}\right)^{\circledast_{-n}}.
\end{split}
\end{equation}

By virtue of low-rank approximation, only very small matrices are involved to perform the multiplications in (\ref{eq:eqlrau}) and (\ref{eq:eqlrag}). In addition, the computational complexities of $(\bm{U}^{(s)^\odot})^T\textrm{vec}(\bm{\mathcal{M}}^{(s)})$ and $\bm{\mathcal M}_{(n)}^{(s)}\bm{U}^{{(s)}^{\odot_{-n}}}$ are respectively reduced to $\mathcal{O}(SR\tilde{R}\sum_{n}\!I_n)$ and $\mathcal{O}(NSR\tilde{R}\sum_{n}\!I_n)$ via the transformation of (\ref{eq:eqlrag}) and (\ref{eq:eqlrau}) (here we set $\tilde R^{(s)}=\tilde{R}$). 

Overall, to develop an efficient coupled tensor decomposition algorithm, we further propose the lraCoNCPD-APG algorithm based on CoNCPD-APG algorithm and low-rank approximation. The implementation of lraCoNCPD-APG algorithm includes two steps: (i) performing unconstrained CPD of tensors $\bm{\mathcal M}^{(s)}$ successively to achieve their low-rank approximation as $\bm{\mathcal M}^{(s)}\approx\left\llbracket \tilde{\bm{U}}^{(1,s)},\tilde{\bm{U}}^{(2,s)},\cdots,\tilde{\bm{U}}^{(N,s)} \right\rrbracket$; (ii) updating $\bm{U}_k^{(n,s)}$ and $\bm{\mathcal D}_k^{(s)}$ via solving the optimization problem in (\ref{eq:cosfunclra}) with fixed $\tilde{\bm{U}}^{(1,s)},\tilde{\bm{U}}^{(2,s)},\cdots,\tilde{\bm{U}}^{(N,s)}$. The framework of lraCoNCPD-APG algorithm is presented in Algorithm \ref{2nd:larCoNCPD-APG}.
\begin{algorithm}[]
      \caption{larCoNCPD-APG algorithm}
      \label{2nd:larCoNCPD-APG}
      \KwIn{$\bm{\mathcal{M}}^{(s)}$, $L_n$, and $R^{(s)}$, $n=1,\cdots,N$, $s=1,\cdots,S$}          
      Initialization:\\ 
$\bm {U}^{(n,s)}$, $\bm{\mathcal D}^{(s)}$, $\bm{\mathcal{M}}_{(n)}^{(s)}$, $n=1,\cdots,N$, $s=1,\cdots,S$\\
Calculate $\tilde{\bm{U}}^{(n,s)}$, $n=1,\cdots,N$, $s=1,\cdots,S$ via unconstrained CPD on $\bm{\mathcal{M}}^{(s)}$, $s=1,\cdots,S$\\
      \For{$k=1,2,\cdots,\textrm{MaxIt}$}
      {
Repeat the 4th to 21st command lines of Algorithm \ref{lst:CoNTD-APG} except the 5th and 10th lines:\\ Calculate $\hat{\bm{\mathcal{G}}}_{k-1}^{(s)}$ and $\hat{\bm G}^{(n,s)}_{k-1}$
via (\ref{eq:solapgG}) and (\ref{eq:solapgU}) by introducing (\ref{eq:eqlrag}) and (\ref{eq:eqlrau})\\
}
     \KwOut{$\bm {U}^{(n,s)}$, $\bm{\mathcal D}^{(s)}$, $n=1,\cdots,N$, $s=1,\cdots,S$}
\end{algorithm}  
\subsection{Remarks and discussion}
\subsubsection{Exact\&flexible coupling}
The coupled NCPD models in (\ref{eq:tensormodel}) consider the exact coupling through a common part which has to be shared by all the tensors $\bm{\mathcal{M}}^{(s)}$ as $\bm{U}_{C}^{(n,1)}=\bm{U}_{C}^{(n,2)}=\cdots=\bm{U}_{C}^{(n,S)}$. Although the coupling assumption is relatively restrictive, the models can also flexibly depict the connections between tensors to some extent. For example, we can impose couplings on any modes or some of them, and the coupling on every mode can be partial or complete. More interestingly, the couplings can be imposed only on specific tensors, and it's easy to derive formulas and modify the code from our work. Please refer to \cite{schenker2020flexible} for more flexibly linear couplings, which will be one of our works.
\subsubsection{Removal of core tensors}
\label{ssub:recoten}
In NCPD, core tensor $\bm{\mathcal{D}}$ is generally merged into the factor matrices that can slightly reduce the computation load. Analogously, we can extend this means to the coupled NCPD, but only for the case where all $N$ modes are not fully coupled among tensors \cite{wang2020group}, i.e., $\exists\, n$ leading to $L_n\!=\!0$. However, for the cases that $\forall\,n$, $L_n\!>\!0$, the core tensors $\bm{\mathcal{D}}^{(s)}$ are required and used to differentiate the magnitude of corresponding components (rank-1 tensor) among tensors. In {Experiment 1}, we design the coupled NCPD problem using synthetic data and verify the importance of core tensors in some cases. In {Experiment 3}, the scaling features of corresponding brain activities provided by core tensors extracted from ERP tensors are used to classify the groups of patients and healthy controls. Thus, the removal of core tensors depends on the coupling constraints on the tensor data.

\begin{table*}[]
\renewcommand\arraystretch{1.2}
\caption{Time complexity per iteration in CoNCPD-APG and \textit{lra}CoNCPD-APG algorithms: \textit{n}th mode of \textit{s}th tensor}
\label{timecomplx}
\centering
\begin{adjustbox}{angle=0}
\begin{threeparttable}
\begin{tabular}{llllll}
    \hline  
		{\bf Equation}&{\bf Operation}&{\bf Description}&{\bf Input size}&{\bf Output size}&{\bf Cost}\\  
    \hline			
		(\ref{eq:solapgG})&$\textcircled{1}=\big(\bm{U}^{{(s)}^T}\bm U^{(s)}\big)^\circledast$&Hadamard product&$I_n\times R,n=1,2,\cdots,N$&$R\times R$&$R^2\sum_{n=1}^N\!{I_n}+R^2N$\\	
		(\ref{eq:solapgG})&$\textcircled{2}=\big(\bm{U}^{(s)^\odot}\big)^T$&Khatri-Rao product&$I_n\times R,n=1,2,\cdots,N$&$R\times \prod_{n=1}^NI_n$&$R\prod_{n=1}^N\!I_n$\\		
		(\ref{eq:solapgG})&$\textcircled{3}=\textcircled{2}\cdot\textrm{vec}\big(\bm{\mathcal{M}}^{(s)}\big)$&Matrix product&$R\times\prod_{n=1}^N\!I_n$, $\prod_{n=1}^N\!I_n\times 1$&$R\times 1$&$R\prod_{n=1}^N\!I_n$\\	
		(\ref{eq:solapgU})&$\textcircled{4}=\big(\bm{U}^{{(s)}^T}\bm U^{(s)}\big)^{\circledast_{-n}}$&Hadamard product&$I_m\times R,m=1,\cdots,N,m\neq n$&$R\times R$&$R^2\sum_{m\neq n}^N\!{I_m}+R^2(N-1)$\\
		(\ref{eq:solapgU})&$\textcircled{5}=\hat{\bm U}^{(n,s)}\cdot\textcircled{4}$&Matrix product&$I_n\times R, R\times R$&$I_n\times R$&$R^2I_n$\\
		(\ref{eq:solapgU})&$\textcircled{6}=\bm{U}^{{(s)}^{\odot_{-n}}}$&Khatri-Rao product&$I_m\times R,m=1,\cdots,N,m\neq n$&$\prod_{m\neq n}^N\!I_m\times R$&$R\prod_{m\neq n}^N\!I_m$\\
		(\ref{eq:solapgU})&$\textcircled{7}=\bm{\mathcal M}_{(n)}^{(s)}\cdot\textcircled{6}$&Matrix product&$I_n\times \prod_{m\neq n}^N\!I_m, \prod_{m\neq n}^N\!I_m\times R$&$I_n\times R$&$R\prod_{n=1}^N\!I_n$\\
		(\ref{eq:eqlrag})&$\textcircled{8}=\big(\bm{U}^{{(s)}^T}\tilde{\bm U}^{(s)}\big)^\circledast$&Hadamard product&$I_n\times R,n=1,2,\cdots,N$&$R\times R$&$R^2\sum_{n=1}^N\!{I_n}+R^2N$\\
		(\ref{eq:eqlrau})&$\textcircled{9}=\left(\tilde{\bm{U}}^{{(s)}^T}\bm U^{(s)}\right)^{\circledast_{-n}}$&Hadamard product&$I_m\times R,m=1,\cdots,N,m\neq n$&$R\times R$&$R^2\sum_{m\neq n}^N\!{I_m}+R^2(N-1)$\\
		(\ref{eq:eqlrau})&$\textcircled{10}=\tilde{\bm{U}}^{(s,n)}\cdot\textcircled{9}$&Matrix product&$I_n\times R, R\times R$&$I_n\times R$&$R^2I_n$\\
    \hline   
\end{tabular}
\begin{tablenotes}
     \item[1] Here let $R^{(s)}=\tilde{R}^{(s)}=R,s=1,2,\cdots,S$.
   \end{tablenotes}
  \end{threeparttable}
     \end{adjustbox}
\end{table*}
\subsubsection{Computational complexity}
For the computational complexity, we mainly calculate the time complexity based on the multiplication operations. From Section \ref{se:section3}, the computational load is mostly dominated by the updates of $\bm{\mathcal D}^{(s)}$ and $\bm{U}^{(n,s)}$, especially the calculation of block-partial gradients $\tilde{\bm{\mathcal{G}}}^{(s)}$ and $\tilde{\bm G}^{(n,s)}$ in (\ref{eq:solapgG}) and (\ref{eq:solapgU}). The costs of calculating them (\textit{n}th mode of \textit{s}th tensor) per iteration are listed in Table \ref{timecomplx}. Let $R^{(s)}\!=\!R$, considering there are $N$ modes and $S$ tensors, the total time complexity for each iteration of CoNCPD-APG algorithm reaches $\mathcal{O}(NSR\prod_{n=1}^N\!I_n)$. By introducing low-rank approximation, (\ref{eq:solapgG}) and (\ref{eq:solapgU}) can be calculated using (\ref{eq:eqlrag}) and (\ref{eq:eqlrau}), the costs of which are also given in Table \ref{timecomplx}. Let $\tilde{R}^{(s)}\!=\!R$, the overall computation load per iteration of lraCoNCPD-APG algorithm is reduced to $\mathcal{O}(NSR^2\sum_{n=1}^N\!In)$. 
\subsubsection{Error bound and convergence analysis}
Inspired by \cite{zhou2012fast}, we also examined the cost and error of low-rank approximation brought for the results of the whole steps. \\
\textbf{Proposition 1:} For a series of tensors $\bm{\mathcal Y}^{(s)}\in\mathbb{R}^{I_1\times I_2\times \cdots I_N}$, suppose that there are nonnegative factor matrices $\bm{U}^{(n,s)}_{\lozenge}$ ($s\leq S,n\leq N$ in this part) satisfying \{$\bm{U}^{(1,s)}_{\lozenge}$,$\bm{U}^{(2,s)}_{\lozenge}$,$\cdots$,$\bm{U}^{(N,s)}_{\lozenge}$\}$=\mathop{\textrm{argmin}}\limits_{\bm{U}^{(n,s)}\geq 0}\sum \limits_{s=1}^S\left \| \bm{\mathcal{Y}}^{(s)} -\left\llbracket \bm{U}^{(1,s)},\bm{U}^{(2,s)},\cdots,\bm{U}^{(N,s)} \right\rrbracket \right \|_{F}$. Let $\left \| \bm{\mathcal{Y}}^{(s)} -\left\llbracket \bm{U}^{(1,s)}_{\lozenge},\bm{U}^{(2,s)}_{\lozenge},\cdots,\bm{U}^{(N,s)}_{\lozenge} \right\rrbracket \right \|_{F}=\epsilon^{(s)}_{\lozenge}$ and $\bm{\mathcal{X}}^{(s)}=$ $\left\llbracket \tilde{\bm{U}}^{(1,s)},\tilde{\bm{U}}^{(2,s)},\cdots,\tilde{\bm{U}}^{(N,s)}\right\rrbracket$ be the low-rank approximation (e.g., unconstrained CPD) to $\bm{\mathcal{Y}}^{(s)}$, then we have\\
(i) 
$
\!\mathop{\textrm{min}}\limits_{\bm{U}^{(\!n\!,\!s)}\!\geq\!0}\!\sum \limits_{\!s\!=\!1\!}^S\!\bigg[\!\left \|\!\bm{\mathcal{Y}}^{\!(\!s\!)\!}\!-\!\bm{\mathcal{X}}^{\!(\!s\!)\!}\!\right\|_F\!+\!\left\|\!\bm{\mathcal{X}}^{\!(\!s\!)\!}\!-\!\left\llbracket\!\bm{U}^{\!(\!1\!,\!s\!)}\!,\!\cdots\!,\!\bm{U}^{\!(\!N\!,\!s\!)}\!\right\rrbracket\!\right \|_{F}\!\bigg]\!=\!\sum \limits_{\!s\!=\!1\!}^S\epsilon_{\!(\!s\!)\!}^{\lozenge}.
$\\
(ii) If $\left \| \bm{\mathcal{Y}}^{(s)}-\bm{\mathcal{X}}^{(s)} \right\|_F=\theta^{(s)}$, suppose that there are nonnegative factor matrices $\vec{\bm{U}}^{(n,s)}$ ($s\leq S,n\leq N$ in this part)  satisfying $\left\lbrace\vec{\bm{U}}^{(1,s)},\vec{\bm{U}}^{(2,s)},\cdots,\vec{\bm{U}}^{(N,s)}\right\rbrace=\mathop{\textrm{argmin}}\limits_{\vec{\bm{U}}^{(n,s)}\geq 0}\sum \limits_{s=1}^S\left \| \bm{\mathcal{X}}^{(s)} -\left\llbracket \bm{U}^{(1,s)},\bm{U}^{(2,s)},\cdots,\bm{U}^{(N,s)} \right\rrbracket \right \|_{F}$. Then $\sum\limits_{s\!=\!1}^S\!\epsilon_{\!(\!s\!)\!}^{\lozenge}\!\leq\!\sum \limits_{\!s\!=\!1\!}^S\!\left \|\!\bm{\mathcal{Y}}^{(\!s\!)}\!-\!\left\llbracket \!\vec{\bm{U}}^{\!(1\!,\!s)}\!,\!\vec{\bm{U}}^{(2\!,\!s)}\!,\!\cdots\!,\!\vec{\bm{U}}^{(N\!,\!s)}\!\right\rrbracket\!\right\|_{F}\!\leq\!\sum \limits_{\!s\!=\!1\!}^S\!\Big[2\theta^{(s)}\!+\!\epsilon_{(s)}^{\lozenge}\Big]$.

The proof of \textbf{Proposition 1} can be found in Appendix \ref{app:appenB}.

Under BCD framework, the objective function of coupled NCPD model in (\ref{eq:costfunc}) was converted into two alternative sets of continuously differentiable functions as in (\ref{eq:eqsolD}) and (\ref{eq:eqsolUns}). Each function is a convex function with Lipchitz continuous gradient, and the solution will be converged to a stationary point as an optimal convergence rate of $\mathcal{O}(\frac{1}{K^2})$, which has been thoroughly demonstrated in \cite{nesterov1983method,guan2012nenmf,xu2013block} and will not be further analyzed in this study.
\subsubsection{Termination criteria}
For the stopping criteria during algorithm optimization, we consider two parameters: the change of average relative error (RelErr) and the maximum number of iterations (MaxIt). We define 
\begin{equation}
\textrm{RelErr}\triangleq\frac{1}{S}\sum_{s=1}^{S} \left[\|\bm{\mathcal{M}}^{(s)}-\bm{\mathcal{\hat{M}}}^{(s)} \|_{F}/ \| \bm{\mathcal{M}}^{(s)} \|_{F}\right]
\end{equation}
where $\bm{\mathcal{M}}^{(s)}$ and $\bm{\mathcal{\hat{M}}}^{(s)}$ are the original and recovered tensors. Furthermore, we stipulate $\left|\textrm{RelErr}_\textrm{new}-\textrm{RelErr}_\textrm{old}\right|<\varepsilon$, i.e., the adjacent RelErr change should be smaller than the preselected threshold. In this study, we choose $\varepsilon=1e-8$ and MaxIt=1000 for the coupled algorithms, and $\varepsilon=1e-4$ and MaxIt=200 for the low-rank approximation.
\subsubsection{Component number selection}
In terms of the number of components and coupled components, i.e., $\tilde{R}^{(s)}$, $R^{(s)}$ and $L_n$, we set ground-truth values for the synthetic experiment. For the real-world data, we adopt the principal component analysis (PCA) method, and the number of principal components with 99\% explained variance is selected as the number of components. Considering that the nonnegative rank is generally no less than the ordinary rank, so we set $\tilde{R}^{(s)}=R^{(s)}$ in this study \cite{zhou2012fast}. Inspired by the tensor spectral clustering method proposed in \cite{hu2022discovering} and the matlab toolbox\footnote{\href{https://github.com/GHu-DUT/Tensor\_Spectral\_Clustering}{https://github.com/GHu-DUT/Tensor\_Spectral\_Clustering}}, we calculate the number of coupled components $L_n$ as: if components from at least half of the tensors have a fairly high correlation coefficient, e.g., 0.7, we consider them as the coupled components. 
\section{Experiments and results}
\label{sec:exp}
In this section, we design and perform three experiments on synthetic data, face image data and two kinds of real-world electroencephalography (EEG) data, aiming to examine and demonstrate the superior performance of CoNCPD-APG and lraCoNCPD-APG algorithms on the coupled NCPD problem. 

\textbf{Experiment settings:} We use alternating least squares (ALS, \cite{cichocki2009nonnegative}) algorithm to implement the low-rank approximation, which has proven to be a reasonable solver for unconstrained CPD problems \cite{zhang2016fast}. The optimization strategies including fast hierarchical alternating least squares (fHALS, \cite{wang2019fast,zdunek2019linked,cichocki2009fast}), multiplicative updating (MU, \cite{lee1999learning,lee2009group}), alternating direction method of multipliers (ADMM, \cite{boyd2011distributed,wang2021shared}) and ALS are used as the competitors to the proposed algorithms. 
For performance comparison, (1) we adopt model fit to evaluate the reconstructed error between original and recovered tensors via $\textrm{Fit}\triangleq \frac{1}{S}\sum_{s}\left[1-\|\bm{\mathcal{M}}^{(s)}-\bm{\mathcal{\hat{M}}}^{(s)} \|_{F}/ \| \bm{\mathcal{M}}^{(s)} \|_{F}\right]$; (2) we record the running time to access the utility of the low-rank approximation; (3) we adopt the performance index (PI) to measure the accuracy of recovered factor matrices via 
\begin{equation}
\textrm{PI}\!\triangleq\!\frac{1}{2\!R\!(\!R\!-\!1\!)\!}\!\left[\! \sum_{i=1}^R\!\bigg(\!\sum_{J=1}^R\!\frac{|g_{ij}|}{\textrm{max}_k|g_{ik}|}\bigg)\!+\!\sum_{i=1}^R\!\bigg(\!\sum_{J=1}^R\!\frac{|g_{ji}|}{\textrm{max}_k|g_{ki}|}\!\bigg)\!\right],    
\end{equation}
where $g_{ij}$ denotes the $(i,j)$th element of the matrix $\bm G = \bm{\tilde{U}}^{\dagger} \bm U$. $\bm{\tilde{U}}$ is the estimation of the factor matrix $\bm U$, and $\dagger$ denotes the pseudo inverse operator. The small value of PI indicates an accurate estimation of the true factor matrix regardless of scale and permutation ambiguity. The input factor matrices and core tensors are initialized with uniformly distributed pseudorandom numbers generated by MATLAB function \texttt{rand}. Signal-to-noise ratio (SNR) is defined as $\textrm{SNR}=10\textrm{log}_{10}(p_s/p_n)$, where $p_s$ and $p_n$ are the signal and noise levels. Then the noise can be added to each tensor as 
\begin{equation}
\bm{\mathcal{M}}_{noise}=p_s\bm{\mathcal{M}}+p_n\frac{\bm{\mathcal{N}}}{\|\bm{\mathcal{N}}\|}\|\bm{\mathcal{M}}\|_F,
\end{equation}
where $\bm{\mathcal{N}}$ is the noise whose entries are drawn from standard normal distribution.

The experiments were carried out with the following computer configurations: CPU-Intel Core i9-14900KF @3.20 GHz; Memory-128 GB; System-64-bit Windows 11; Software-MATLAB R2022b.

\subsection{Experiment 1 Synthetic data}
In this experiment, we design detailed experiments to illustrate the performance of CoNCPD-APG and lraCoNCPD-APG algorithms in terms of decomposition efficiency and accuracy using synthetic data. First, we simulated 20 third-order tensors that are partially/fully coupled in one, two, or three modes according to equation (\ref{eq:tensormodel}). Second, we compare the performance using different levels of noise and tensor sizes. Last, we design some special cases to compare the performance of proposed algorithms with and without core tensors. 
For each experiment, the sizes of tensors are given as $I_{1,2,3}=[100,100,100]$ (except for Exp.~1e, which has varying tensor sizes). The rank of tensors is set to 10, meaning that the ground-truth non-negative factor matrices have 10 components, whose entries are drawn from the uniform distribution. For the noise, we apply a fixed SNR of 20 dB, except for Exp.~1d, which has multiple noise levels. To ensure the reliability of the algorithms, we perform 100 independent runs for each simulation and report the average results.
\paragraph{Experiment 1a: coupled in three modes} 
First, we set the number of coupled components as $L_{1,2,3}=[10,10,10]$ and $[5,5,5]$, corresponding to the fully coupled or partially coupled cases.
\paragraph{Experiment 1b: coupled in two modes}
Then, we consider the coupled cases in two modes, and the number of coupled components among tensors is set to $L_{1,2,3}=[10,10,0]$ and $[5,5,0]$. 
\paragraph{Experiment 1c: coupled in one mode}
Here we examine the coupled cases in one mode, where the number of coupled components among tensors is specified as $L_{1,2,3}=[10,0,0]$ and $[5,0,0]$. 

For the first three simulation experiments, from \tablename~\ref{tab:syndata}, we can see that all algorithms perform almost identically on the tensor fitting values for various coupled types. From the perspective of PI values and running time, except for the second coupling type in Experiment 1a, the proposed CoNCPD-APG algorithm is very competitive in terms of both decomposition speed and accuracy of the recovered factor matrix. Although the ALS-based algorithm runs faster than others, its decomposition accuracy for factor matrices is slightly lower, indicating that the convergence is not thorough enough. Conversely, while the ADMM-based algorithm converges slowly, it achieves high decomposition accuracy. For the proposed lraCoNCPD-APG algorithm, both the unconstrained ALS and APG optimization steps converge rapidly, achieving high tensor fitting and accurate estimation of the factor matrices. Compared to methods without low-rank approximation, our approach achieves convergence 5 to 10 times faster with only a minimal loss in decomposition precision.
\begin{table}[]
\caption{Performance comparison of the algorithms for tensor models that are coupled in one, two, and three modes optimized using different algorithms (The values in the row * are the performance of two steps in the lracoNCPD-APG algorithm)}
\label{tab:syndata}
\renewcommand{\arraystretch}{1.2}
\setlength{\tabcolsep}{3pt}
\centering
%\begin{threeparttable}
\begin{tabular}{llcccccc}
\toprule
\multicolumn{2}{c}{Settings}& \multicolumn{2}{c}{Experiment 1a}&\multicolumn{2}{c}{Experiment 1b}&\multicolumn{2}{c}{Experiment 1c}\cr
\cmidrule(lr){3-4}
\cmidrule(lr){5-6}
\cmidrule(lr){7-8}
\multicolumn{2}{c}{$L_{1,2,3}$}&[10,10,10]&[5,5,5]&[10,10,0]&[5,5,0]&[10,0,0]&[5,0,0]\cr
\midrule
\multirow{7}{*}{Fit}
&ADMM    &0.9534&0.9528&0.9536&0.9536&0.9535&0.9536\cr
&MU      &0.9534&0.9484&0.9526&0.9532&0.9525&0.9527\cr
&ALS     &0.9507&0.9534&0.9534&0.9535&0.9517&0.9534\cr
&fHALS   &0.9534&0.9531&0.9536&0.9536&0.9533&0.9535\cr
&APG     &0.9534&0.9524&0.9536&0.9536&0.9531&0.9535\cr
&lra\&APG  &0.9522&0.9492&0.9529&0.9526&0.9522&0.9521\cr
%&*step1,2&0.9488,\;0.9799&0.9488,\;0.9852&0.9515,\;0.9886&0.9517,\;0.9909&0.9514,\;0.9887&0.9515,\;0.9914\cr
&*step1&0.9488&0.9488&0.9515&0.9517&0.9514&0.9515\cr
&*step2&0.9799&0.9852&0.9886&0.9909&0.9887&0.9914\cr
\cmidrule(lr){2-8}
\multirow{7}{*}{PI}
&ADMM    &0.0040&0.0185&0.0039&0.0043&0.0041&0.0042\cr
&ALS     &0.0087&0.0069&0.0045&0.0049&0.0110&0.0059\cr
&MU      &0.0040&0.0593&0.0183&0.0093&0.0188&0.0147\cr
&fHALS   &0.0039&0.0092&0.0045&0.0054&0.0055&0.0059\cr
&APG     &0.0039&0.0199&0.0040&0.0043&0.0055&0.0042\cr
&lra\&APG  &0.0079&0.0500&0.0113&0.0124&0.0152&0.0185\cr
\cmidrule(lr){2-8}
\multirow{7}{*}{Time}
&ADMM    &15.993&19.387&9.1603&9.3587&9.9832&9.2803\cr
&ALS     &5.8274&5.5828&2.9349&2.4715&3.8710&2.8685\cr
&MU      &51.409&38.017&19.834&22.979&20.789&20.298\cr
&fHALS   &6.8604&12.200&7.7722&3.4584&8.5071&6.7957\cr
&APG     &8.5072&18.440&5.2630&5.6902&5.6701&5.9041\cr
&lra\&APG  &1.4356&2.1844&1.0896&1.0934&1.1867&1.2285\cr
%&*step1,2&0.7959,\;0.6397&0.6381,\;1.5463&0.5381,\;0.5515&0.5360,\;0.5574&0.5757,\;0.6110&0.5688,\;0.6598\cr
&*step1&0.7959&0.6381&0.5381&0.5360&0.5757&0.5688\cr
&*step2&0.6397&1.5463&0.5515&0.5574&0.6110&0.6598\cr
\bottomrule
\end{tabular}
%\end{threeparttable}
\end{table}
\paragraph{Experiment 1d: different noises}
In this scenario, we demonstrate and compare the performance of the proposed algorithms across a range of SNR values (from 0 to 20dB with steps of 2dB) under the following settings: partially coupled in two modes as $L_{1,2,3}=[5,5,0]$. 

\begin{figure}[t]
\centering
\includegraphics[width=8cm,height=4cm]{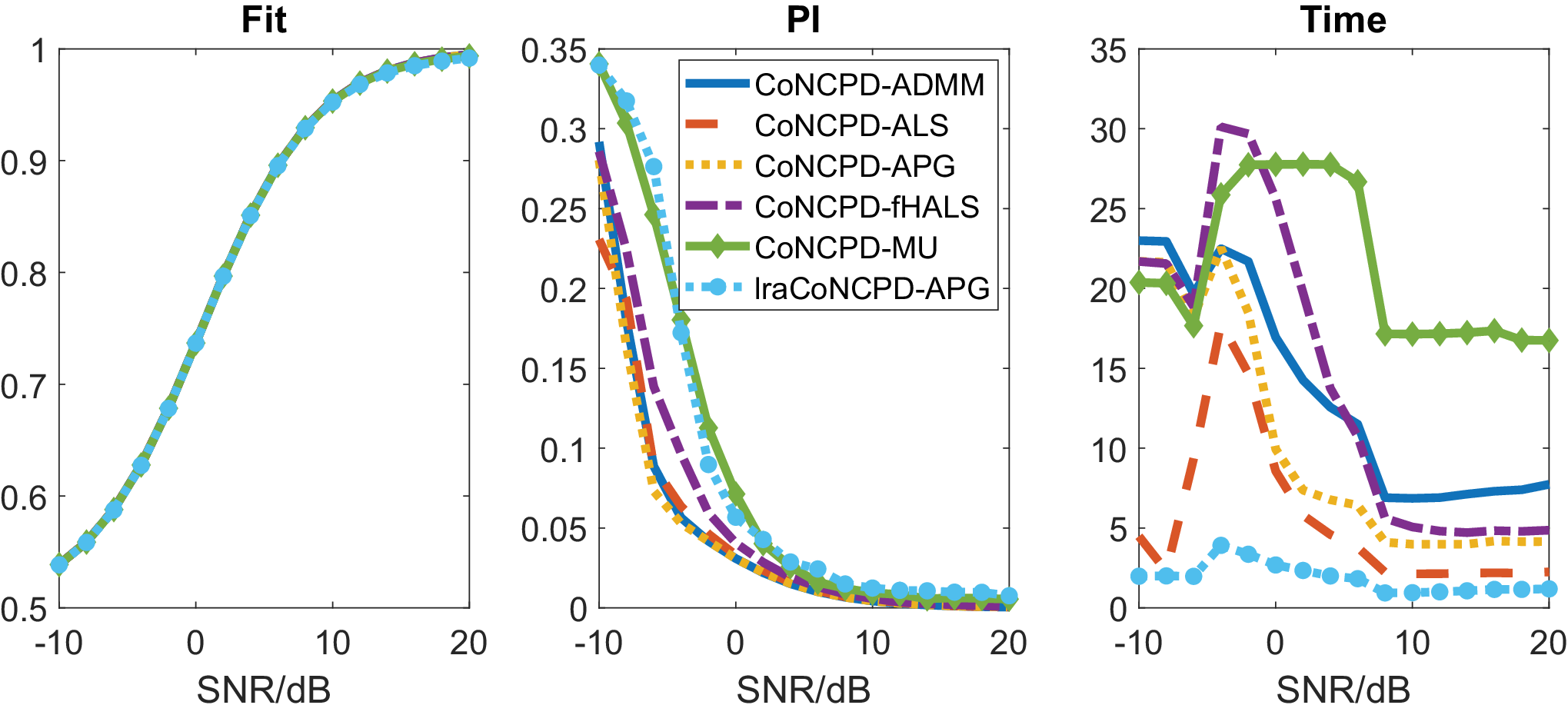}
\caption{Performance comparison of CoNCPD-APG, lraCoNCPD-APG, and their competitors at different additive noise levels}
\label{fig:expe1d}
\end{figure}

As can be seen from \figurename~\ref{fig:expe1d}, all algorithms exhibit very similar performance in terms of tensor fit across the range of SNR values. Regarding PI values, all algorithms perform comparably at high SNRs. However, the CoNCPD-APG algorithm outperforms the others at low SNRs, while ranking in the middle in terms of running time. The introduction of LRA greatly improves the running speed, though it slightly compromises decomposition accuracy.
\paragraph{Experiment 1e: different sizes}
To evaluate the decomposition performance with varying tensor sizes, we define the tensor dimensions as $I_{1,2,3}=[10n,10n,10n]$, where $n$ ranges from 1 to 20 with steps of 2. We adopt the case with partially coupled data across two modes $L_{1,2,3}=[5,5,0]$.

\begin{figure}[t]
\centering
\includegraphics[width=8cm,height=4cm]{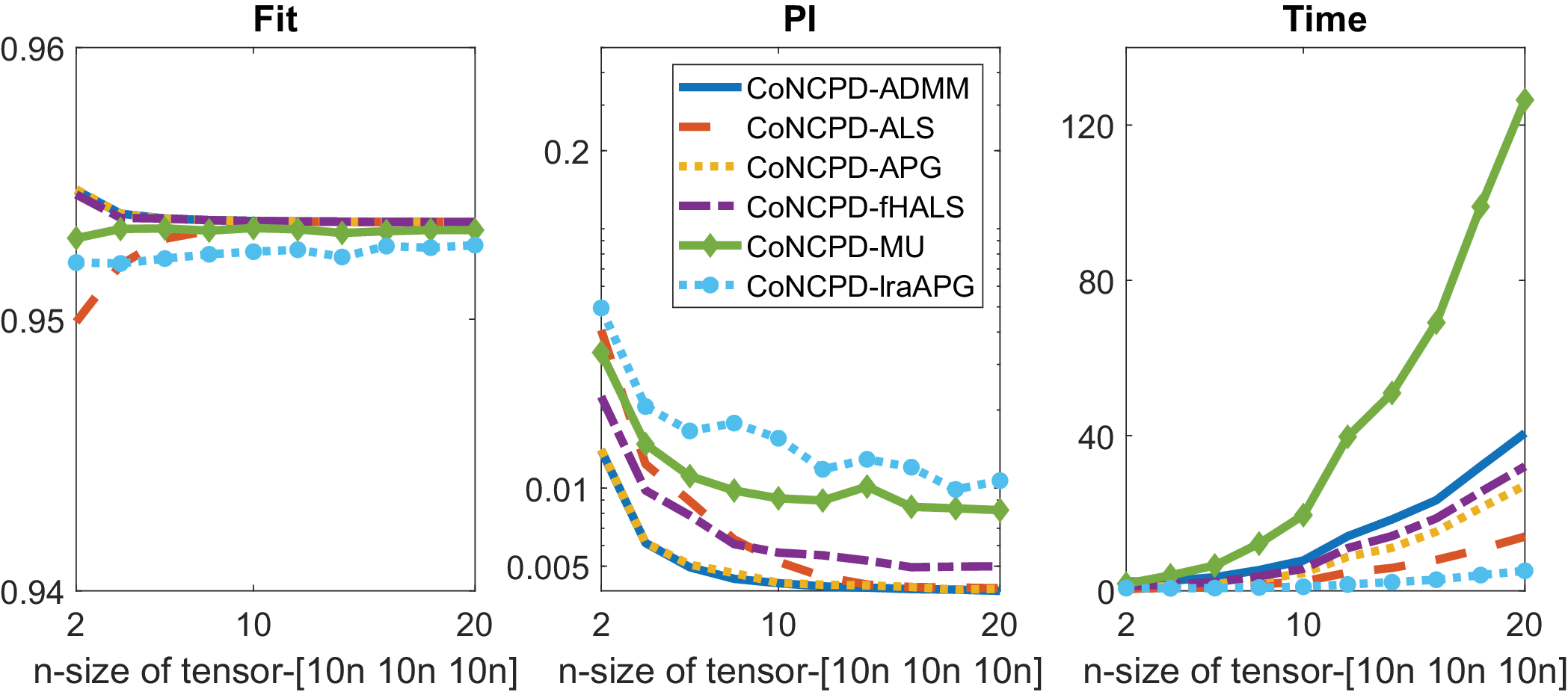}
\caption{Performance comparison of CoNCPD-APG, lraCoNCPD-APG, and their competitors using different tensor sizes}
\label{fig:expe1e}
\end{figure}

In this case, the change in tensor size has little effect on the tensor fit, but as the tensor size increases, the PI value becomes smaller and the running time becomes significantly longer, as shown in \figurename~\ref{fig:expe1e}. The CoNCPD-APG and CoNCPD-ADMM algorithms have similar performance in terms of PI values, while CoNCPD-APG takes less time to run. With LRA, lraCoNCPD-APG can greatly reduce the execution time, and this advantage becomes more significant as the size of tensors increases. Meanwhile, its cost
is only a slight reduction in decomposition accuracy.
\paragraph{Experiment 1f: core tensors or not}
As discussed in \ref{ssub:recoten}, the core tensors are necessary for the cases that $\forall\,n$, $L_n\!>\!0$. Here we take the case of $L_{1,2,3}=[5,5,5]$ as an example, and the algorithms without optimizing core tensors are respectively termed as CoNCPD-APG-NC and lraCoNCPD-APG-NC.

\begin{figure}[t]
\centering
\includegraphics[width=8cm,height=4cm]{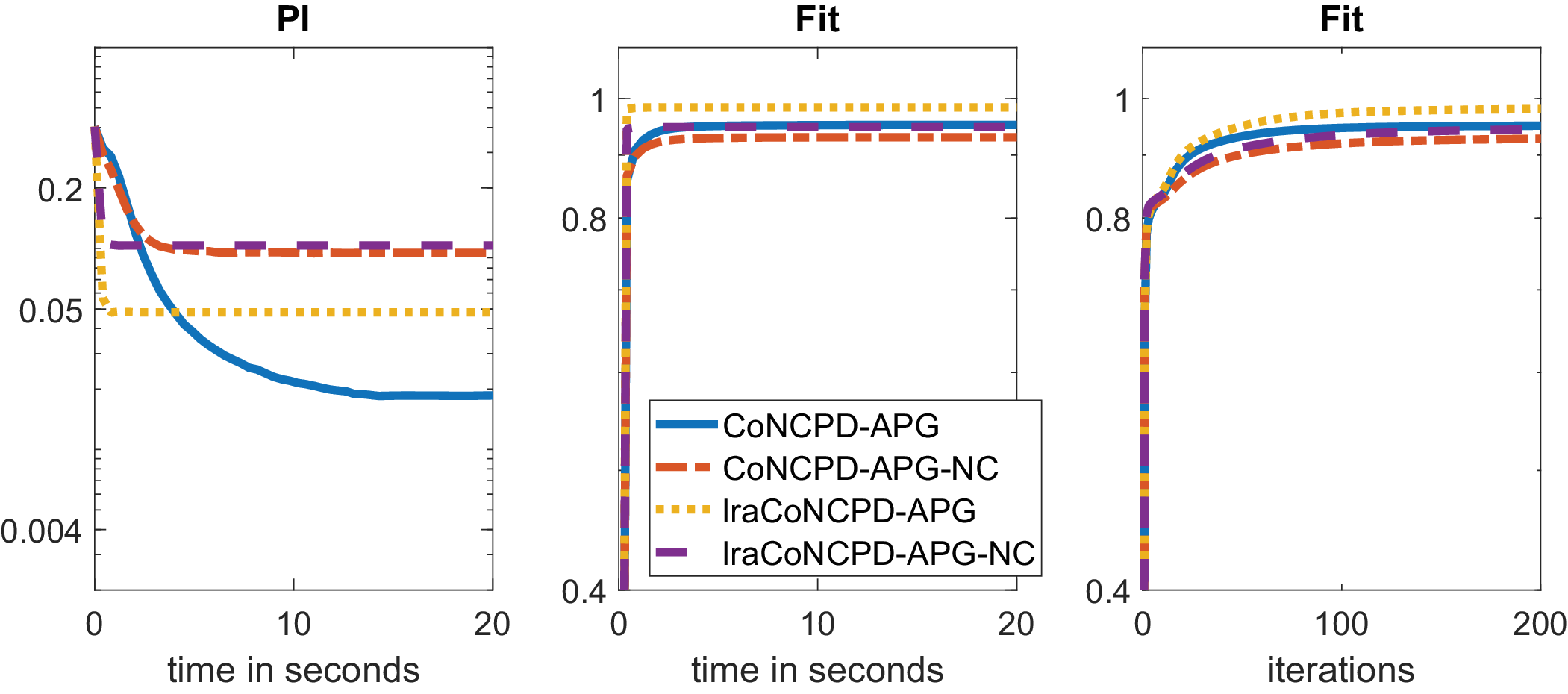}
\caption{Performance comparison of CoNCPD-APG, lraCoNCPD-APG, CoNCPD-APG-NC, and lraCoNCPD-APG-NC algorithms.}
\label{fig:expe1f}
\end{figure}

From \figurename~\ref{fig:expe1f}, we can see that CoNCPD-APG achieves the best decomposition accuracy on PI values, followed by lraCoNCPD-APG, CoNCPD-APG-NC and lraCoNCPD-APG-NC. It should be noted that the tensor fitting curve of lraCoNCPD-based algrothms is drawn from its second step, which is why it is higher than CoNCPD-based algorithms. This is only to illustrate that the convergence accuracy of the second step of the algorithm is still relatively high. Actually, for lraCoNCPD-APG algorithm, the final global tensor fit is 0.9, which is lower than 0.95 of CoNCPD-APG, followed by CoNCPD-APG-NC with 0.85 and lraCoNCPD-APG-NC with 0.8. To sum up, the core tensors are required in some cases.
%\begin{figure*}
%\centering
%\includegraphics[width=14cm,height=12cm]{./imgs/SyntheticDataResults.eps}
%\caption{PI, Tenfit, Time and ObjFun curves of all compared algorithms versus the size of tensors under %SNR=20dB and 50 independent runs.}
%\label{fig:synthedata}
%\end{figure*}
%From Figure~\ref{fig:synthedata}, we can see that CoNCPD-APG achieves the best decomposition accuracy in terms of PI, Tenfit and ObjFun, followed by lraCoNCPD-APG, CoNCPD-APG-NC and lraCoNCPD-APG-NC. However, CoNCPD-APG is the most time-consuming, followed by CoNCPD-APG-NC and then by lraCoNCPD-APG and lraCoNCPD-APG-NC. The introduction of low-rank approximation can greatly reduce the execution time, and this advantage becomes more significant as the size of tensors increases. Meanwhile, its cost is only a slight reduction in decomposition accuracy. Since the core tensors are not updated in the optimization process, the time consumption of CoNCPD-APG-NC is also alleviated, but the decomposition accuracy of CoNCPD-APG-NC and lraCoNCPD-APG-NC is reduced to some extent.

\subsection{Experiment 2 Face image data} 
In this experiment, we conduct coupled tensor decomposition analysis using the extended Yale B face database\footnote{\href{http://vision.ucsd.edu/~leekc/ExtYaleDatabase/ExtYaleB.html}{http://vision.ucsd.edu/~leekc/ExtYaleDatabase/ExtYaleB.html}} for image reconstruction and denoising. This database contains gray-scale face images of 38 subjects acquired from 9 poses and 64 illumination conditions, here we only use the cropped images under frontal pose of all illuminations\footnote{\href{http://www.cad.zju.edu.cn/home/dengcai/Data/FaceData.html}{http://www.cad.zju.edu.cn/home/dengcai/Data/FaceData.html}}\cite{georghiades2001few,cai2007spectral}. In this case, each subject corresponds to 64 images, and each image is cropped and resized to 32$\times$32 pixels. Finally, we have 31 third-order tensors with the size of 32 pixels $\times$32 pixels $\times$64 condition by stacking face images of each subject along illumination conditions (7 subjects are removed due to data incompleteness). Using the PCA-based method, the component number of tensors for 31 subjects are separately selected as \{24, 23, 23, 24, 23, 24, 24, 23, 22, 24, 23, 24, 22, 23, 23, 24, 23, 24, 23, 23, 26, 23, 23, 24, 23, 23, 23, 24, 25, 22, 23, 22, 22, 23, 24, 24, 25, 23\}. Specifically, through unfolding along the first mode, each tensor is converted into a matrix with the size of 32$\times$2048. Then performing the PCA on the matrices successively, and the number of principal components with a total explained variance exceeds 99\% is regarded as the corresponding component number.  
Using the tensor spectral clustering method \cite{hu2022discovering}, we select those components that have a correlation value greater than 0.7 and from at least 70\% of the tensors as coupled components, and finally 15 coupled components among tensors are extracted. 
\begin{table}[]
\caption{Performance comparison of the algorithms in the image reconstruction and denoising based on coupled NCPD model}
\label{tab:YaleBPerfComp}
\renewcommand{\arraystretch}{1.2}
\centering
%\begin{threeparttable}
\begin{tabular}{lcccccc}
\toprule
\multirow{2}{*}{Settings}&\multicolumn{3}{c}{Noise=0.1}&\multicolumn{3}{c}{Noise=0.0001}\cr
\cmidrule(lr){2-4}
\cmidrule(lr){5-7}
&Fit&Time&PSNR&Fit&Time&PSNR\cr
\midrule
ADMM    &0.5053&37.015&21.134&0.7688&36.662&22.066\cr
ALS     &0.4940&30.652&20.149&0.7193&34.510&20.354\cr
MU      &0.5064&26.062&21.324&0.7773&26.634&22.413\cr
fHALS   &0.5028&31.358&20.954&0.7583&29.535&21.691\cr
APG     &0.5079&32.830&21.480&0.7816&30.308&22.596\cr
lra\&APG&0.5061&8.708&21.396&0.7794&8.6834&22.504\cr
*step1&0.4938&0.5285&-&0.8353&1.0323&-\cr
*step2&0.8266&8.2146&-&0.8464&7.6512&-\cr
\bottomrule
\end{tabular}
%\end{threeparttable}
\end{table}

The performance of the proposed algorithms is compared with their competitors at two noise levels: 0.1 and 0.0001 salt-and-pepper noise. Here we use the matlab function $\texttt{imnoise}$ with the parameter \texttt{salt \& pepper} to add the noise. Peak-signal-to-noise ratio (PSNR) is also used to evaluate the quality of reconstructed face images. Table \ref{tab:YaleBPerfComp} gives the algorithm performance comparison averaged from 100 independent runs. We can see that the proposed APG-based algorithms perform slightly better than other algorithms in tensor fittings and PSNR, indicating high decomposition accuracy and stronger image reconstruction capabilities. More importantly, by introducing low-rank approximation strategy, the lraCoNCPD-APG algorithm can greatly reduce execution time while maintaining decomposition performance. 

\begin{figure*}
\centering
\includegraphics[width=16cm,height=10cm]{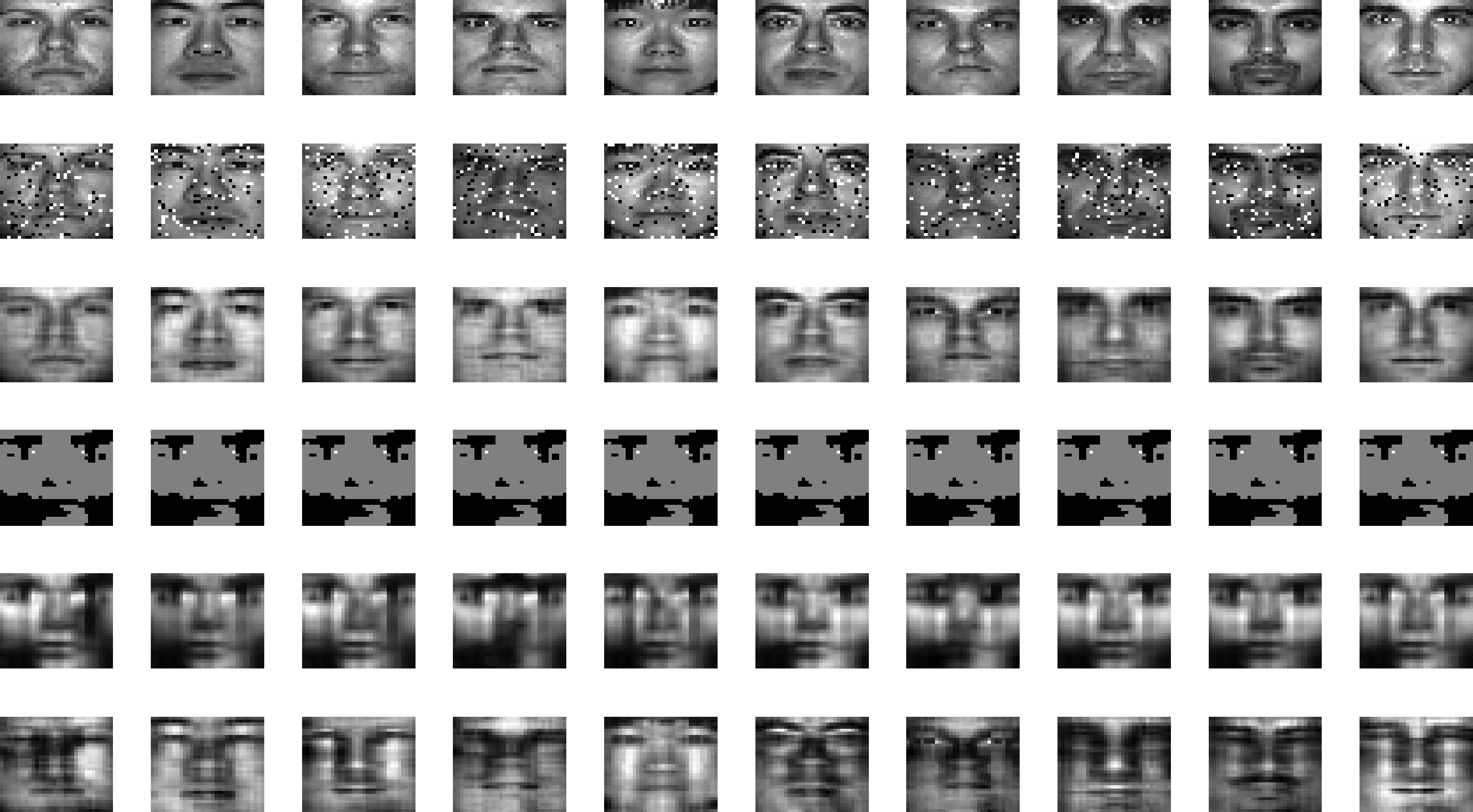}
\caption{Illustration of original (1st row), noisy (2nd row), and recovered (3rd row) face images, as well as common (4th and 5th rows) and individual (6th row) face image features obtained by CoNCPD-APG algorithm for image reconstruction and denoising with 0.1 salt-and-pepper noise}
\label{fig:Faceimages}
\end{figure*}

The execution time of lra-based algorithm showed in the table includes the running time of the unconstrained CPD and APG optimization, and we also give the performance of each separate step. In addition, Figure \ref{fig:Faceimages} gives an illustration of original, noisy, and recovered face images, as well as common and individual face image features obtained by CoNCPD-APG algorithm. The fourth row of images are only common features among different images, while the fifth row is the common features multiplied by the values from core tensors, and the last row is the individual features corresponding to the particular images.

\subsection{Experiment 3 Real-world ERP data} 
In this experiment, we compare the proposed CoNCPD-APG and lraCoNCPD-APG algorithms with CoNCPD-MU, CoNCPD-fHALS and CoNCPD-MU algorithms in the multi-domain feature extraction of event-related potential (ERP) data \footnote{\href{http://www.escience.cn/people/cong/AdvancedSP\_ERP.html}{http://www.escience.cn/people/cong/AdvancedSP\_ERP.html}}. Two groups of data are chosen: 21 children with reading disability (RD) and 21 children with attention deficit (AD), aiming to acquire multi-domain features of ERP data which can better discriminate the two groups. Using complex Morlet wavelet transform, we generate the third-order tensors of 42 subjects (21 RD \& 21 AD) with the size of 9 (channels) $\times$ 71 (frequency bins) $\times$ 60 (temporal points) to testify the effectiveness and practicality of coupled tensor decomposition. Following \cite{cong2012benefits}, we set the number of components to $R^{(1)}=R^{(2)}=\cdots R^{(42)}=36$. Considering the nature of ERP data, we assume that these third-order ERP tensors are coupled in spatial, spectral and temporal modes, and we directly set the number of coupled components to 36. 
\begin{figure*}
\centering
\includegraphics[width=16cm,height=3.4cm]{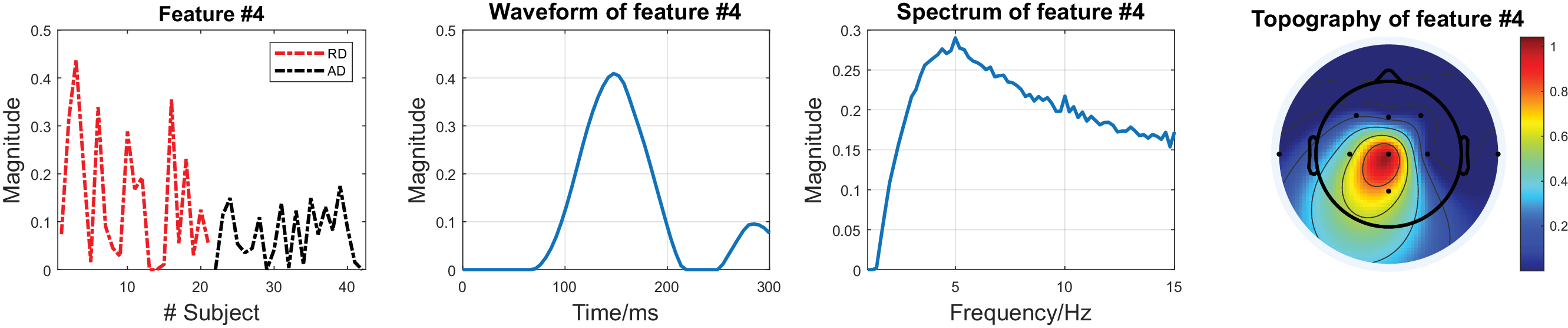}
\caption{An example of multi-domain feature and its related temporal, spectral and spatial components of ERP data extracted by CoNCPD-APG algorithm}
\label{fig:ERPMultiDoFeaplot}
\end{figure*}

ERP data are acquired through repeated presentation of stimuli, which makes their properties in temporal, spectral and spatial domains roughly known before they are actually extracted. According to prior knowledge given in \cite{cong2012benefits}, we can select the expected multi-domain features and their corresponding temporal, spectral and spatial components from the decomposition results of ERP data. Figure \ref{fig:ERPMultiDoFeaplot} gives an example illustration of multi-domain features and their corresponding components extracted by CoNCPD-APG algorithm in the 1st run. For multi-domain feature shown
in the figure, statistical analysis using t-test reveals the significant difference between RD and AD groups with $t_{20} = 2.0899$, $p = 0.0108$. The relevant temporal component (latency peaks around 150 ms) and spectral component (spectrum peaks around 5 Hz) closely match the property of mismatch negativity component \cite{cong2012benefits}. The corresponding topography denotes that the difference of RD and AD groups may appear in the central and left hemisphere \cite{cong2012benefits}.

We adopt three steps to verify the stability of multi-domain feature extraction of all the compared algorithms in 100 runs. (1) We select the multi-domain features and their parallel three components in the 1st runs of 5 algorithms. (2) We average the selected ones separately as a set of template patterns, which are termed as $\bm{u}^{\textrm{fea}}_{\textrm{temp}}$, $\bm{u}^{\textrm{tem}}_{\textrm{temp}}$, $\bm{u}^{\textrm{spe}}_{\textrm{temp}}$ and $\bm{u}^{\textrm{spa}}_{\textrm{temp}}$. (3) We define the maximum correlation coefficient (MCC) between template patterns and feature-based components of $k$th runs as follows
\begin{equation}
\begin{split}
\textrm{MCC}(k)=\textrm{max}\left[\textrm{corr}(\bm{u}^{\textrm{fea}}_{\textrm{temp}}, \bm{U}^{\textrm{fea}}_k)\circledast\textrm{corr}(\bm{u}^{\textrm{tem}}_{\textrm{temp}}, \bm{U}^{\textrm{tem}}_k)\right.\\\left.\circledast\textrm{corr}(\bm{u}^{\textrm{spe}}_{\textrm{temp}}, \bm{U}^{\textrm{spe}}_k)\circledast\textrm{corr}(\bm{u}^{\textrm{spa}}_{\textrm{temp}}, \bm{U}^{\textrm{spa}}_k)\right]
\end{split}
\end{equation}
where $k$ denotes the run number and \texttt{corr} is a matlab function which returns a vector containing the pairwise linear correlation coefficient between $\bm u$ and $\bm U$. $\bm{U}^{\textrm{fea}}_k$, $\bm{U}^{\textrm{tem}}_k$, $\bm{U}^{\textrm{spe}}_k$ and $\bm{U}^{\textrm{spa}}_k$ represent multi-domain features and their corresponding temporal, spectral and spatial components in the $k$th run, respectively. Obviously, if the MCC is close to 1, it means that the extraction of multi-domain features is more stable.
\begin{table}
\caption{Performance comparison of the algorithms in multi-domain feature extraction of ERP data based on coupled NCPD model}
\label{tab:ERPPerfComp}
\renewcommand{\arraystretch}{1.2}
\setlength{\tabcolsep}{1.2pt}
\centering
%\begin{threeparttable}
\begin{tabular}{lcccccccc}
\toprule
Methods&ADMM&ALS&MU&fHALS&APG&lra\&APG&*step-1&*step-2\cr
\midrule
Fit&0.8222&0.7639&0.8375&0.8350&0.8491&0.8490&0.9901&0.8493\cr
Time&63.392&69.123&54.962&62.978&56.050&12.317&0.9768&11.340\cr
MCC-Mean&0.7143&0.6318&0.7936&0.8567&0.8828&0.8777&-&-\cr
MCC-SD&0.1471&0.1681&0.1155&0.1118&0.0796&0.0799&-&-\cr
\bottomrule
\end{tabular}
%\end{threeparttable}
\end{table} 

Table \ref{tab:ERPPerfComp} gives the average tensor fit and running time from 100 independent runs for ERP data, as well as the means and SDs of MCCs for multi-domain features. We can see that the proposed CoNCPD-APG and lraCoNCPD-APG algorithms are superior to competitors in terms of decomposition accuracy and multi-domain feature extraction stability. Interestingly, although some algorithms achieve good performance in data fittings, their multi-domain feature extraction is less stable. This experiment also demonstrates the positive effect of low-rank approximation in large-scale tensor coupling analysis, we can see that the first step of the algorithm takes very little time, but saves a lot of time for the second step.
\subsection{Real-world Ongoing EEG data}
In this part, we apply the proposed algorithms to the joint analysis of ongoing EEG data, which are collected from 14 participants when listening to an 8.5-min piece of modern tango. After the short-time Fourier transform and necessary preprocessing, we eventually generate 14 third-order tensors with the size of 46 (frequency bins) $\times$ 510 (time samples) $times$ 64 (space channels) for 14 participants. Detailed information about data acquisition and preprocessing can be found in \cite{cong2013linking, wang2020group}. For ease of comparison, following \cite{wang2020group}, which assumes that coupling information exists in both spatial and spectral modes, we set the number of components as the number of components were respectively selected as \{44, 34, 36, 38, 36, 39, 35, 35, 34, 37, 33, 36, 34, 35\}, and set the number of component components as  $L_{1,2,3}=[26,26,0]$. Next, we follow the flowchart of ongoing EEG data processing and analysis proposed in \cite{wang2020group}, including  data acquisition \& preprocessing, musical feature extraction, tensor representation, algorithm implementation, correlation analysis, hierarchical clustering, and cluster selection of interest. 

\begin{figure*}
\centering
\includegraphics[width=16cm,height=9cm]{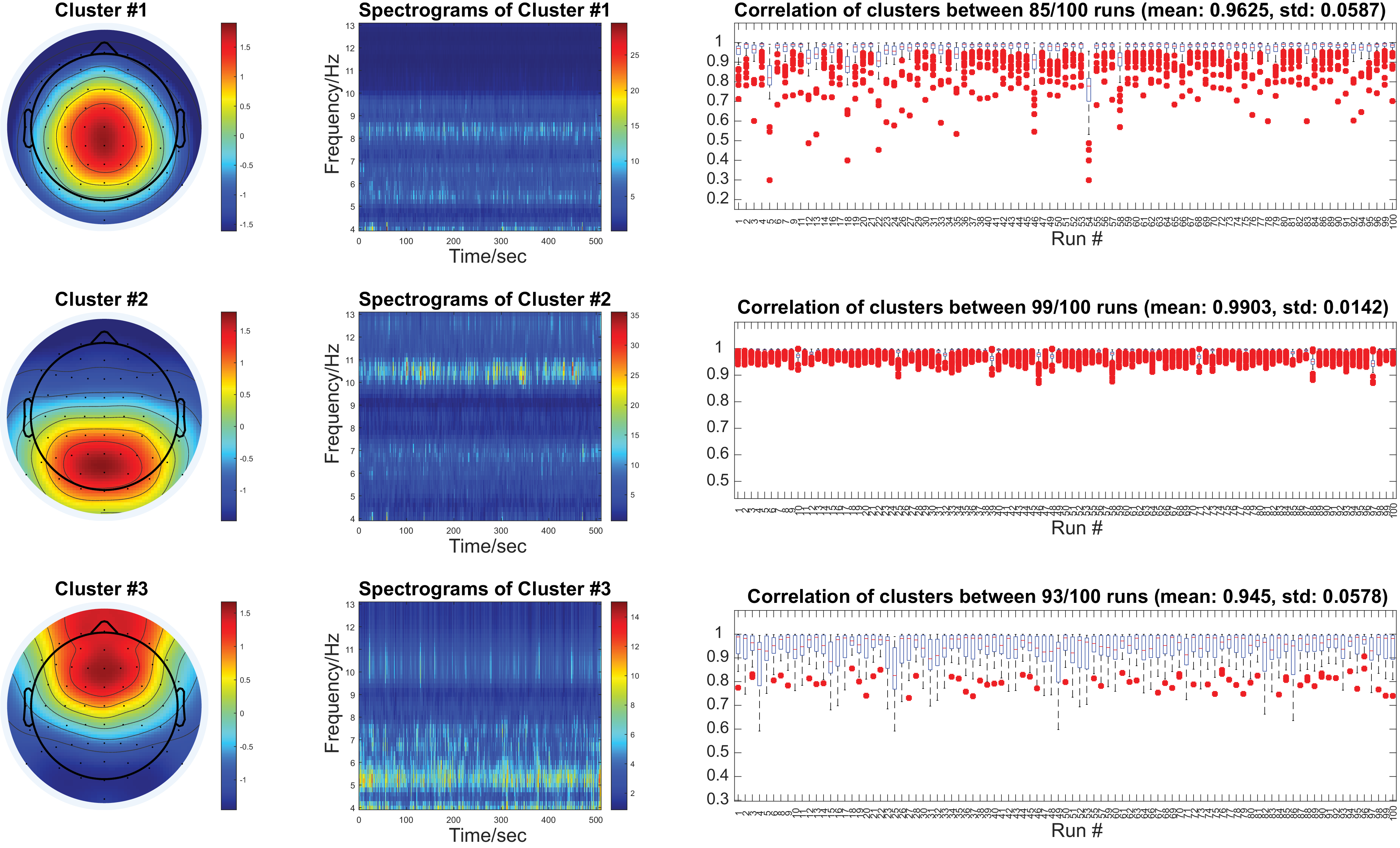}
\caption{Illustrations (Topography, spectrogram, and correlation) of clusters \#I, \#II and \#III obtained from 100 runs by the CoNCPD-APG algorithm}
\label{fig:experiment4_APG_plo}
\end{figure*}

Finally, we conclude three clusters of interest, where the topographies reveal activations in centro-parietal, occipito-parietal, and frontal regions, as well as oscillations mainly in theta, alpha, and theta-alpha frequency bands, see \figurename~\ref{fig:experiment4_APG_plo}, and these findings are consistent with the previous study \cite{wang2020group}. Moreover, the CoNCPD-APG algorithm shows more stable performance in extracting clusters of interest, as the probability of clusters \#I, \#II, and \#III appearing in 100 runs reaches 85\%, 99\%, and 93\%, while the probability of clusters \#I, \#II, and \#III obtained by the lraCoNCPD-APG algorithm is 75\%, 94\%, and 89\%, respectively. Meanwhile, the average inter-run correlations of clusters \#I, \#II, and \#III of the CoNCPD-APG algorithm reached 0.9625, 0.9903, and 0.945. Although the performance of the lraCoNCPD-APG algorithm declined, it also reached 0.9573, 0.9859, and 0.9521, respectively. In addition, in terms of convergence, the CoNCPD-APG algorithm performs slightly better than the lraCoNCPD-APG algorithm in tensor fitting (0.7219 vs 0.7143), but requires more running time (49.27s vs 7.86s).

\section{Conclusion}
In this study, we addressed the problem of coupled tensor decomposition, focusing on the simultaneous decomposition of nonnegative multi-block tensors. To enhance convergence speed and optimization accuracy, we first proposed a coupled nonnegative CANDECOMP/PARAFAC decomposition algorithm based on the alternating proximal gradient (APG) method, referred to as CoNCPD-APG. Building on this, we further proposed the lraCoNCPD-APG algorithm combining APG and low-rank approximation. We also gave some discussions on the properties of these algorithms as well as some implementation remarks. Experiments of synthetic data, real-world face image data, and event-related potential (ERP) data were conducted to compare the proposed algorithms with fast hierarchical alternating least squares (fHALS), multiplicative updating (MU), and alternating least squares (ALS)-based algorithms in the context of coupled NCPD problems. Results illustrated that our proposed algorithms are superior to competitors in terms of decomposition accuracy, image reconstruction capability, and multi-domain feature extraction stability, and also demonstrated that the introduction of low-rank approximation can greatly improve the computation efficiency without compromising the decomposition quality.

Determining the number of coupled components depends on the validity of potential assumptions and relevant prior knowledge. So far, its selection in real-world applications remains somewhat subjective, making it an open issue and a subject for our future research. Furthermore,  exploring linearly soft coupling will also be a focus of our future work.

% Note that the IEEE does not put floats in the very first column
% - or typically anywhere on the first page for that matter. Also,
% in-text middle ("here") positioning is typically not used, but it
% is allowed and encouraged for Computer Society conferences (but
% not Computer Society journals). Most IEEE journals/conferences use
% top floats exclusively. 
% Note that, LaTeX2e, unlike IEEE journals/conferences, places
% footnotes above bottom floats. This can be corrected via the
% \fnbelowfloat command of the stfloats package.
\section*{Acknowledgment}
This work is supported by National Natural Science Foundation of China (Grant No.91748105), National Foundation in China (No. JCKY2019110B009 \& 2020-JCJQ-JJ-252), the Fundamental Research Funds for the Central Universities [DUT20LAB303\& DUT20LAB308] in Dalian University of Technology in China, Dalian Science and Technology Talent Innovation Support Project (No.2023RY034), and Dalian life and health guidance program project (No.2024033). This study is to memorize Prof. Tapani Ristaniemi for his great help to the authors, Fengyu Cong and Xiulin Wang.

% or


\appendices
\section{Algorithm derivations and parameter settings}
\label{app:appenA}
For completeness of this paper, in this section, some steps of algorithm derivations and relevant parameter settings will be further explained below.
\subsubsection{\bf{CoNCPD-APG algorithm}}
Following \cite{guan2012nenmf,xu2013block}, by using Lagrange multiplier method, we obtain (\ref{eq:solD}), (\ref{eq:solUns}), (\ref{eq:solUnC}) and (\ref{eq:solUnsI}) from (\ref{eq:eqsolD}) and (\ref{eq:eqsolUns}). When updating the core tensor $\bm{\mathcal{D}}^{(s)}$, by keeping all factor matrices $\bm{U}^{(n,s)}$ fixed, we first convert (\ref{eq:costfunc}) to
\begin{equation}
\label{eq:costfuncG}
F_d=\frac{1}{2}\left\|\textrm{vec}\big(\bm{\mathcal M}^{(s)}\big)-\bm{U}^{(s)\odot}\textrm{ddiag}\big({\bm{\mathcal{D}}}^{(s)}\big)\right \|^{2}_F
\end{equation}
where $\textrm{vec}(\bm{\mathcal M}^{(s)})$ denotes the vectorization of tensor $\bm{\mathcal M}^{(s)}$. Mathematically, the squared Frobenius norm of a matrix can be replaced by the trace of multiplication of the matrix and its transpose. Then (\ref{eq:costfuncG}) can be represented as:
\begin{equation}
\begin{split}
\label{eq:costfuncGG}
F_d=\frac{1}{2}\textrm{tr}\left[ \left( \textrm{vec}\big(\bm{\mathcal M}^{(s)}\big)-\bm{U}^{(s)\odot}\textrm{ddiag}\big({\bm{\mathcal{D}}}^{(s)}\big)\right)^T\right.\\\left.\left( \textrm{vec}\big(\bm{\mathcal M}^{(s)}\big)-\bm{U}^{(s)\odot}\textrm{ddiag}\big({\bm{\mathcal{D}}}^{(s)}\big)\right)\right]
\end{split}
\end{equation}

According to trace property, the block-partial gradient $\hat{\bm{\mathcal G}}^{(s)}$ of (\ref{eq:costfuncGG}) with respect to $\hat{\bm{\mathcal D}}^{(s)}$ can be calculated by
\begin{equation}
\begin{split}
\label{eq:solbpgG}
\hat{\bm{\mathcal G}}^{(s)}&=\nabla_{\hat{\bm {\mathcal D}}^{(s)}}F_d \\&=\textrm{ddiag}\left[\left(\bm{U}^{(s)\odot}\right)^T\bm{U}^{(s)\odot}\textrm{ddiag}\left(\hat {\bm {\mathcal D}}^{(s)}\right)\right]\\&-\textrm{ddiag}\left[\left(\bm{U}^{(s)\odot}\right)^T\textrm{vec}\left(\bm{\mathcal{M}}^{(s)}\right)\right]
\end{split}
\end{equation}
where the outer notation `ddiag' means the tensorization from a vector to a super-diagonal tensor, which is the reverse operation of inner one. Using the property of Khatri-Rao product, we can efficiently calculate $(\bm{U}^{(s)\odot})^T\bm{U}^{(s)\odot}$ by
\begin{equation}
\left(\bm{U}^{(s)\odot}\right)^T\bm{U}^{(s)\odot}=\left(\bm{U}^{{(s)}^T}\bm U^{(s)}\right)^\circledast
\end{equation}

When updating the factor matrix (without coupling information) $\bm{U}^{(n,s)}$, by keeping all other variables $\bm{U}^{(m,s)}, m\neq n$ and $\bm{\mathcal{D}}^{(s)}$ fixed, (\ref{eq:costfunc}) is represented as follows:
\begin{equation}
\label{eq:costfuncU}
F_u=\frac{1}{2}\left \| \bm{\mathcal{M}}_{(n)}^{(s)} -\bm{U}^{(n,s)}\bm{D}^{(s)}\left(\bm{U}^{(s)\odot_{-n}}\right)^T \right \|_{F}^{2}
\end{equation}
where $\bm{\mathcal{M}}_{(n)}^{(s)}$ denotes the mode-$n$ matricization of $\bm{\mathcal{M}}^{(s)}$. $\bm D^{(s)}$ is a diagonal matrix and its diagonal elements correspond to the super-diagonal elements of core tensor $\bm {\mathcal{D}}^{(s)}$. Similarly, the block-partial gradient $\hat{\bm G}^{(n,s)}$ of (\ref{eq:costfuncU}) at $\hat{\bm U}^{(n,s)}$ can be calculated as:
\begin{equation}
\begin{split}
\label{eq:solbpgU}
\hat{\bm G}^{(n,s)}&=\nabla_{\hat{\bm U}^{(n,s)}} F_u\\&=\hat{\bm U}^{(n,s)}\bm D^{(s)}\left(\bm{U}^{(s)\odot_{-n}}\right)^T\bm{U}^{{(s)}\odot_{-n}}\left(\bm D^{(s)}\right)^T\\&-\bm{\mathcal M}_{(n)}^{(s)}\bm{U}^{{(s)}\odot_{-n}}\left(\bm D^{(s)}\right)^T\\&=\hat{\bm U}^{(n,s)}\bm D^{(s)}\left(\bm{U}^{{(s)}^T}\bm U^{(s)}\right)^{\circledast_{-n}}\bm D^{{(s)}}\\&-\bm{\mathcal M}_{(n)}^{(s)}\bm{U}^{{(s)}\odot_{-n}}\bm D^{{(s)}}
\end{split}
\end{equation}

However, when updating the factor matrix $\bm{\mathit{U}}^{(n,s)}$ which consists of two parts: $\bm{U}_{C}^{(n,s)}$ and $\bm{U}_{I}^{(n,s)}$, we first substitute $\bm{U}^{(n,s)}=\left [\bm{U}_{C}^{(n,s)}\;\bm{U}_{I}^{(n,s)}\right]$ into (\ref{eq:costfuncU}) and have
\begin{equation}
\label{eq:costfuncUCI}
F_{ci}\!=\!\frac{1}{2}\left \| \bm{\mathcal{M}}_{(n)}^{(s)}\!-\!\left[\bm{U}_C^{(n,s)}\,\bm{U}_I^{(n,s)}\right]\!\bm{D}^{(s)}\!\left(\bm{U}^{(s)\odot_{-n}}\right)^T\right \|_{F}^{2}
\end{equation}
Let $\bm B^{(n,s)}\!=\!\bm{D}^{(s)}\left(\bm{U}^{(s)\odot_{-n}}\right)^T\in\mathbb{R}_+^{R^{(s)}\times \prod_{m\neq n}^{N}I_{m}}$. Let $\hat{\bm G}^{(n,s)}_C$ denote the block-partial gradient of (\ref{eq:costfuncUCI}) at $\hat{\bm U}_{C}^{(n,s)}$, which can be calculated as:
\begin{equation}
\begin{split}
\label{eq:solbpgUCI}
\hat{\bm G}^{(n,s)}_C&=\nabla_{\hat{\bm U}_{C}^{(n,s)}}F_{ci}\\&=\hat{\bm U}_{C}^{(n,s)}\bm B_C^{(n,s)}\left(\bm B_C^{(n,s)}\right)^T\\&-\left(\bm{\mathcal{M}}_{(n)}^{(s)}-\hat{\bm U}_I^{(n,s)}\bm B_I^{(n,s)}\right)\left(\bm B_C^{(n,s)}\right)^T
\\&=\left(\hat{\bm U}^{(n,s)}\bm B^{(n,s)}-\bm{\mathcal{M}}_{(n)}^{(s)}\right)\left(\bm B_C^{(n,s)}\right)^T
\end{split}
\end{equation}
where $\bm{\hat U}_C^{(n,s)}$ denotes an extrapolated point of $\bm{U}_C^{(n,s)}$. $\bm B^{(n,s)}=\left [\bm B^{(n,s)}_C;\bm B^{(n,s)}_I\right]$, $\bm B_C^{(n,s)}\in\mathbb{R}_+^{L_n\times \prod_{m\neq n}^{N}I_{m}}$ and $\bm B_I^{(n,s)}\in\mathbb{R}_+^{(R^{(s)}-L_n)\times \prod_{m\neq n}^{N}I_{m}}$. From (\ref{eq:solbpgU}) and (\ref{eq:solbpgUCI}), it can be infered that $\hat{\bm G}^{(n,s)}_C$ is equal to $\hat{\bm G}^{(n,s)}_{:,1:L_n}$. Analogously, the block-partial gradient $\hat{\bm G}^{(n,s)}_I$ at $\hat{\bm U}_{I}^{(n,s)}$ can be obtained as $\hat{\bm G}^{(n,s)}_{:,L_n:R^{(s)}}$ and $\bm{\hat G}^{(n,s)}=\left[\bm{\hat G}_C^{(n,s)}\;\bm{\hat G}_I^{(n,s)}\right]$.
\subsubsection{\bf{Parameter settings}}
Consider updating $\bm{\mathcal D}^{(s)}$ and $\bm{U}^{(n,s)}$ at the $k$th iteration. Following [-], we set the Lipschitz constants $L_{d,k-1}^{(s)}$ and $L_{u,k-1}^{(n,s)}$ as:
\begin{equation}
\label{eq:eqLipsLd}
L_{d,k-1}^{(s)} = \left\| (\bm{U}_{k-1}^{(s)\odot})^T\bm{U}_{k-1}^{(s)\odot}\right \|
\end{equation}
and 
\begin{equation}
\label{eq:eqLipsLu}
L_{u,k-1}^{(n,s)}=\left\| \bm D_{k-1}^{(s)} (\bm{U}_{k-1}^{(s)\odot_{-n}})^T\bm{U}_{k-1}^{(s)\odot_{-n}}\bm D_{k-1}^{(s)}\right \|
\end{equation}
where $\|\cdot\|$ denotes the spectral norm. Using the property of Khatri-Rao product, we can also efficiently calculate $L_{d,k-1}^{(s)}$ and $L_{u,k-1}^{(n,s)}$ through
\begin{equation}
\begin{split}
&\left(\bm{U}_{k-1}^{(s)\odot}\right)^T\bm{U}_{k-1}^{(s)\odot}=\left[\left(\bm{U}_{k-1}^{(n,s)}\right)^T\bm U_{k-1}^{(n,s)}\right]^\circledast\\&
\left(\bm{U}_{k-1}^{(s)\odot_{-n}}\right)^T\bm{U}_{k-1}^{{(s)}\odot_{-n}}=\left[\left(\bm{U}_{k-1}^{(m,s)}\right)^T\bm U_{k-1}^{(m,s)}\right]^{\circledast_{-n}}
\end{split}
\end{equation}

We take the extrapolation weights as
\begin{equation}
\label{eq:eqextrawd}
w_{d,k-1}^{(s)}=\textrm{min}\left( \hat w_{k-1},\; \delta_w\sqrt{\frac{L_{d,k-2}^{(s)}}{L_{d,k-1}^{(s)}}}\right)
\end{equation}
and 
\begin{equation}
\label{eq:eqextrawu}
w_{u,k-1}^{(n,s)}=\textrm{min}\left( \hat w_{k-1},\; \delta_w\sqrt{\frac{L_{u,k-2}^{(n,s)}}{L_{u,k-1}^{(n,s)}}}\right)
\end{equation}
where $\delta_w<1$ is predefined (e.g., 0.9999, \cite{xu2015alternating}), and $\hat w_{k-1}=\frac{t_{k-1}-1}{t_k}$ with $t_0=1$ and $t_k=\frac{1}{2}\left(1+\sqrt{1+4t^2_{k-1}}\right)$. Moreover, we define the extrapolation at points ${\bm {\mathcal D}}^{(s)}_{k-1}$ and ${\bm U}^{(n,s)}_{k-1}$ as
\begin{equation}
\label{eq:eqextraPGG}
\hat{\bm {\mathcal D}}^{(s)}_{k-1}={\bm {\mathcal D}}^{(s)}_{k-1}+w^{(s)}_{d,k-1}\left({\bm{\mathcal D}}^{(s)}_{k-1}-{\bm{\mathcal {D}}}^{(s)}_{k-2}\right)
\end{equation}
and 
\begin{equation}
\label{eq:eqextraPUU}
\hat{\bm U}^{(n,s)}_{k-1}={\bm U}^{(n,s)}_{k-1}+w^{(n,s)}_{u,k-1}\left({\bm U}^{(n,s)}_{k-1}-{\bm U}^{(n,s)}_{k-2}\right)
\end{equation}

\section{Proof of \textbf{Proposition 1}}
\label{app:appenB}
(i) Let $\zeta(\bm{\mathcal{X}}^{(s)},\bm{U}^{(n,s)})|_{{\bm{U}^{(n,s)}\geq 0,\bm{\mathcal{X}}^{(s)}\in\mathbb{R}^{I_1\times I_2\times \cdots I_N}}}=$ $\sum\limits_{s=1}^{S}\bigg[\left\|\bm{\mathcal{Y}}^{(s)}\!-\!\bm{\mathcal{X}}^{(s)}\right\|_F\!+\!\left\|\bm{\mathcal{X}}^{(s)}\!-\!\left\llbracket \bm{U}^{(1,s)}\!,\!\bm{U}^{(2,s)}\!,\!\cdots\!,\!\bm{U}^{(N,s)} \right\rrbracket\right\|_F\bigg]$. Therefore, for $\forall\; \bm{U}^{(n,s)}\geq 0$ and $\bm{\mathcal{X}}^{(s)}\in\mathbb{R}^{I_1\times I_2\times \cdots I_N}$, we have
\begin{equation}
\begin{split}
&\sum\limits_{s=1}^{S}\!\bigg[\!\left\|\!\bm{\mathcal{Y}}^{(s)}\!-\!\bm{\mathcal{X}}^{(s)}\right\|_F\!+\!\left\|\bm{\mathcal{X}}^{(s)}\!-\!\left\llbracket \bm{U}^{(1,s)}\!,\!\bm{U}^{(2,s)}\!,\!\cdots\!,\!\bm{U}^{(N,s)} \right\rrbracket\!\right\|_F\!\bigg]\\&
\geq \sum\limits_{s=1}^{S}\bigg[\left\|\bm{\mathcal{Y}}^{(s)}\!-\!\left\llbracket \bm{U}^{(1,s)}\!,\!\bm{U}^{(2,s)}\!,\!\cdots\!,\!\bm{U}^{(N,s)} \right\rrbracket\right\|_F\bigg]\\&
\geq \sum\limits_{s=1}^{S}\bigg[\left\|\bm{\mathcal{Y}}^{(s)}\!-\!\left\llbracket \bm{U}^{(1,s)}_{\lozenge}\!,\!\bm{U}^{(2,s)}_{\lozenge}\!,\!\cdots\!,\!\bm{U}^{(N,s)}_{\lozenge} \right\rrbracket\right\|_F\bigg]\\&
=\sum\limits_{s=1}^{S}\epsilon_{(s)}^{\lozenge}
\end{split}
\end{equation}
Obviously, $\zeta$ can reach the lower bound if $\bm{U}^{(n,s)}=\bm{U}^{(n,s)}_{\lozenge}$ and $\bm{\mathcal{X}}^{(s)}=\left\llbracket \bm{U}^{(1,s)}_{\lozenge}\!,\!\bm{U}^{(2,s)}_{\lozenge}\!,\!\cdots\!,\!\bm{U}^{(N,s)}_{\lozenge} \right\rrbracket$. This completes the proof of (i).\\
(ii) According to the definition of $\epsilon_{(s)}^{\lozenge}$, first we can have 
\begin{equation}
\sum \limits_{s=1}^S\epsilon_{(s)}^{\lozenge}\leq \sum \limits_{s=1}^S\left \| \bm{\mathcal{Y}}^{(s)} -\left\llbracket \vec{\bm{U}}^{(1,s)},\vec{\bm{U}}^{(2,s)},\cdots,\vec{\bm{U}}^{(N,s)} \right\rrbracket \right \|_{F}
\end{equation}
Then we have 
\begin{equation}
\begin{split}
&\sum \limits_{s=1}^S\left \| \bm{\mathcal{Y}}^{(s)} -\left\llbracket \vec{\bm{U}}^{(1,s)},\vec{\bm{U}}^{(2,s)},\cdots,\vec{\bm{U}}^{(N,s)} \right\rrbracket \right \|_{F}\\&
\!\leq\!\sum \limits_{s=1}^S\!\bigg[\!\left\|\!\bm{\mathcal{Y}}^{(s)}\!-\!\bm{\mathcal{X}}^{(s)}\!\right\|_F\!+\!\left \| \!\bm{\mathcal{X}}^{(s)}\!-\!\left\llbracket \vec{\bm{U}}^{(1,s)}\!,\!\vec{\bm{U}}^{(2,s)}\!,\!\cdots\!,\!\vec{\bm{U}}^{(N,s)}\! \right\rrbracket \right\|_{F}\!\bigg]\\&
\!=\!\sum \limits_{s=1}^S\!\bigg[\!\theta^{(s)}\!+\!\left \| \!\bm{\mathcal{X}}^{(s)}\!-\!\left\llbracket \vec{\bm{U}}^{(1,s)}\!,\!\vec{\bm{U}}^{(2,s)}\!,\!\cdots\!,\!\vec{\bm{U}}^{(N,s)}\! \right\rrbracket \right\|_{F}\!\bigg]\\&
\!\leq\!\sum \limits_{s=1}^S\!\bigg[\!\theta^{(s)}\!+\!\left \| \!\bm{\mathcal{X}}^{(s)}\!-\!\left\llbracket \bm{U}^{(1,s)}_{\lozenge}\!,\!\bm{U}^{(2,s)}_{\lozenge}\!,\!\cdots\!,\!\bm{U}^{(N,s)}_{\lozenge}\! \right\rrbracket \right\|_{F}\!\bigg]\\&
\!\leq\!\sum \limits_{s=1}^S\!\bigg[\!\theta^{(s)}\!+\left\|\!\bm{\mathcal{X}}^{(s)}\!-\!\bm{\mathcal{Y}}^{(s)}\!\right\|_F\!+\!\left \| \!\bm{\mathcal{Y}}^{(s)}\!-\!\left\llbracket \bm{U}^{\!(\!1\!,\!s\!)\!}_{\lozenge}\!,\!\bm{U}^{\!(\!2\!,\!s\!)}_{\lozenge}\!,\!\cdots\!,\!\bm{U}^{\!(\!N\!,\!s\!)}_{\lozenge}\! \right\rrbracket \right\|_{F}\!\bigg]\\&
\!\leq\!\sum \limits_{s=1}^S\bigg[2\theta^{(s)}+\epsilon_{(s)}^{\lozenge}\bigg]
\end{split}
\end{equation}
This completes the proof of (ii).
%\section{Additional figure in Experiment 4}
%\begin{figure*}[h]
%\centering
%\includegraphics[width=16cm,height=9cm]{./imgs/Fig_experiment4_lraAPG_plot.eps}
%\caption{Illustrations (topography, spectrogram, and correlation) of clusters \#I, \#II and \#III obtained from 100 runs by the lraCoNCPD-APG algorithm}
%\label{fig:experiment4_lraAPG_plo}
%\end{figure*}

\ifCLASSOPTIONcaptionsoff
  \newpage
\fi

\bibliographystyle{IEEEtran}
\balance
\bibliography{refs}

\begin{thebibliography}{10}
\providecommand{\url}[1]{#1}
\csname url@samestyle\endcsname
\providecommand{\newblock}{\relax}
\providecommand{\bibinfo}[2]{#2}
\providecommand{\BIBentrySTDinterwordspacing}{\spaceskip=0pt\relax}
\providecommand{\BIBentryALTinterwordstretchfactor}{4}
\providecommand{\BIBentryALTinterwordspacing}{\spaceskip=\fontdimen2\font plus
\BIBentryALTinterwordstretchfactor\fontdimen3\font minus
  \fontdimen4\font\relax}
\providecommand{\BIBforeignlanguage}[2]{{%
\expandafter\ifx\csname l@#1\endcsname\relax
\typeout{** WARNING: IEEEtran.bst: No hyphenation pattern has been}%
\typeout{** loaded for the language `#1'. Using the pattern for}%
\typeout{** the default language instead.}%
\else
\language=\csname l@#1\endcsname
\fi
#2}}
\providecommand{\BIBdecl}{\relax}
\BIBdecl

\bibitem{hitchcock1927expression}
F.~L. Hitchcock, ``The expression of a tensor or a polyadic as a sum of
  products,'' \emph{Journal of Mathematics and Physics}, vol.~6, no. 1-4, pp.
  164--189, 1927.

\bibitem{harshman1970foundations}
R.~A. Harshman \emph{et~al.}, ``Foundations of the parafac procedure: Models
  and conditions for an" explanatory" multimodal factor analysis,'' 1970.

\bibitem{carroll1970analysis}
J.~D. Carroll and J.-J. Chang, ``Analysis of individual differences in
  multidimensional scaling via an n-way generalization of “eckart-young”
  decomposition,'' \emph{Psychometrika}, vol.~35, no.~3, pp. 283--319, 1970.

\bibitem{cichocki2009nonnegative}
A.~Cichocki, R.~Zdunek, A.~H. Phan, and S.-i. Amari, \emph{Nonnegative matrix
  and tensor factorizations: applications to exploratory multi-way data
  analysis and blind source separation}.\hskip 1em plus 0.5em minus 0.4em\relax
  John Wiley \& Sons, 2009.

\bibitem{cong2015tensor}
F.~Cong, Q.-H. Lin, L.-D. Kuang, X.-F. Gong, P.~Astikainen, and T.~Ristaniemi,
  ``Tensor decomposition of eeg signals: a brief review,'' \emph{Journal of
  neuroscience methods}, vol. 248, pp. 59--69, 2015.

\bibitem{cong2013multi}
F.~Cong, A.-H. Phan, P.~Astikainen, Q.~Zhao, Q.~Wu, J.~K. Hietanen,
  T.~Ristaniemi, and A.~Cichocki, ``Multi-domain feature extraction for small
  event-related potentials through nonnegative multi-way array decomposition
  from low dense array eeg,'' \emph{International journal of neural systems},
  vol.~23, no.~02, p. 1350006, 2013.

\bibitem{kolda2009tensor}
T.~G. Kolda and B.~W. Bader, ``Tensor decompositions and applications,''
  \emph{SIAM review}, vol.~51, no.~3, pp. 455--500, 2009.

\bibitem{cichocki2015tensor}
A.~Cichocki, D.~Mandic, L.~De~Lathauwer, G.~Zhou, Q.~Zhao, C.~Caiafa, and H.~A.
  Phan, ``Tensor decompositions for signal processing applications: From
  two-way to multiway component analysis,'' \emph{IEEE signal processing
  magazine}, vol.~32, no.~2, pp. 145--163, 2015.

\bibitem{sidiropoulos2017tensor}
N.~D. Sidiropoulos, L.~De~Lathauwer, X.~Fu, K.~Huang, E.~E. Papalexakis, and
  C.~Faloutsos, ``Tensor decomposition for signal processing and machine
  learning,'' \emph{IEEE Transactions on Signal Processing}, vol.~65, no.~13,
  pp. 3551--3582, 2017.

\bibitem{zhou2016linked}
G.~Zhou, Q.~Zhao, Y.~Zhang, T.~Adal{\i}, S.~Xie, and A.~Cichocki, ``Linked
  component analysis from matrices to high-order tensors: Applications to
  biomedical data,'' \emph{Proceedings of the IEEE}, vol. 104, no.~2, pp.
  310--331, 2016.

\bibitem{wang2020group}
X.~Wang, W.~Liu, P.~Toiviainen, T.~Ristaniemi, and F.~Cong, ``Group analysis of
  ongoing eeg data based on fast double-coupled nonnegative tensor
  decomposition,'' \emph{Journal of neuroscience methods}, vol. 330, p. 108502,
  2020.

\bibitem{jonmohamadi2019extraction}
Y.~Jonmohamadi, S.~Muthukumaraswamy, J.~Chen, J.~Roberts, R.~Crawford, and
  A.~Pandey, ``Extraction of common task features in eeg-fmri data using
  coupled tensor-tensor decomposition,'' \emph{bioRxiv}, p. 685941, 2019.

\bibitem{gong2018double}
X.~Gong, Q.~Lin, F.~Cong, and L.~De~Lathauwer, ``Double coupled canonical
  polyadic decomposition for joint blind source separation,'' \emph{IEEE
  Transactions on Signal Processing}, vol.~66, no.~13, pp. 3475--3490, 2018.

\bibitem{sorensen2015coupled}
M.~S{\o}rensen, I.~Domanov, and L.~De~Lathauwer, ``Coupled canonical polyadic
  decompositions and (coupled) decompositions in multilinear
  rank-(l\_r,n,l\_r,n,1) terms---part ii: Algorithms,'' \emph{SIAM Journal on
  Matrix Analysis and Applications}, vol.~36, no.~3, pp. 1015--1045, 2015.

\bibitem{hunyadi2017tensor}
B.~Hunyadi, P.~Dupont, W.~Van~Paesschen, and S.~Van~Huffel, ``Tensor
  decompositions and data fusion in epileptic electroencephalography and
  functional magnetic resonance imaging data,'' \emph{Wiley Interdisciplinary
  Reviews: Data Mining and Knowledge Discovery}, vol.~7, no.~1, p. e1197, 2017.

\bibitem{xue2019coupled}
Z.~Xue, S.~Yang, H.~Zhang, and P.~Du, ``Coupled higher-order tensor
  factorization for hyperspectral and lidar data fusion and classification,''
  \emph{Remote Sensing}, vol.~11, no.~17, p. 1959, 2019.

\bibitem{cichocki2013tensor}
A.~Cichocki, ``Tensor decompositions: a new concept in brain data analysis?''
  \emph{arXiv preprint arXiv:1305.0395}, 2013.

\bibitem{calhoun2009review}
V.~D. Calhoun, J.~Liu, and T.~Adal{\i}, ``A review of group ica for fmri data
  and ica for joint inference of imaging, genetic, and erp data,''
  \emph{Neuroimage}, vol.~45, no.~1, pp. S163--S172, 2009.

\bibitem{gong2015generalized}
X.-F. Gong, X.-L. Wang, and Q.-H. Lin, ``Generalized non-orthogonal joint
  diagonalization with lu decomposition and successive rotations,'' \emph{IEEE
  Transactions on Signal Processing}, vol.~63, no.~5, pp. 1322--1334, 2015.

\bibitem{morup2011applications}
M.~M{\o}rup, ``Applications of tensor (multiway array) factorizations and
  decompositions in data mining,'' \emph{Wiley Interdisciplinary Reviews: Data
  Mining and Knowledge Discovery}, vol.~1, no.~1, pp. 24--40, 2011.

\bibitem{yokota2012linked}
T.~Yokota, A.~Cichocki, and Y.~Yamashita, ``Linked parafac/cp tensor
  decomposition and its fast implementation for multi-block tensor analysis,''
  in \emph{International Conference on Neural Information Processing}.\hskip
  1em plus 0.5em minus 0.4em\relax Springer, 2012, pp. 84--91.

\bibitem{wang2019generalization}
X.~Wang, C.~Zhang, T.~Ristaniemi, and F.~Cong, ``Generalization of linked
  canonical polyadic tensor decomposition for group analysis,'' in
  \emph{International Symposium on Neural Networks}.\hskip 1em plus 0.5em minus
  0.4em\relax Springer, 2019, pp. 180--189.

\bibitem{zhou2015group}
G.~Zhou, A.~Cichocki, Y.~Zhang, and D.~P. Mandic, ``Group component analysis
  for multiblock data: Common and individual feature extraction,'' \emph{IEEE
  transactions on neural networks and learning systems}, vol.~27, no.~11, pp.
  2426--2439, 2015.

\bibitem{acar2011all}
E.~Acar, T.~G. Kolda, and D.~M. Dunlavy, ``All-at-once optimization for coupled
  matrix and tensor factorizations,'' \emph{arXiv preprint arXiv:1105.3422},
  2011.

\bibitem{acar2017acmtf}
E.~Acar, Y.~Levin-Schwartz, V.~D. Calhoun, and T.~Adali, ``Acmtf for fusion of
  multi-modal neuroimaging data and identification of biomarkers,'' in
  \emph{2017 25th European Signal Processing Conference (EUSIPCO)}.\hskip 1em
  plus 0.5em minus 0.4em\relax IEEE, 2017, pp. 643--647.

\bibitem{chatzichristos2018fusion}
C.~Chatzichristos, M.~Davies, J.~Escudero, E.~Kofidis, and S.~Theodoridis,
  ``Fusion of eeg and fmri via soft coupled tensor decompositions,'' in
  \emph{2018 26th European Signal Processing Conference (EUSIPCO)}.\hskip 1em
  plus 0.5em minus 0.4em\relax IEEE, 2018, pp. 56--60.

\bibitem{schenker2023parafac2}
C.~Schenker, X.~Wang, and E.~Acar, ``Parafac2-based coupled matrix and tensor
  factorizations,'' in \emph{ICASSP 2023-2023 IEEE International Conference on
  Acoustics, Speech and Signal Processing (ICASSP)}.\hskip 1em plus 0.5em minus
  0.4em\relax IEEE, 2023, pp. 1--5.

\bibitem{wang2021shared}
X.~Wang, W.~Liu, X.~Wang, Z.~Mu, J.~Xu, Y.~Chang, Q.~Zhang, J.~Wu, and F.~Cong,
  ``Shared and unshared feature extraction in major depression during music
  listening using constrained tensor factorization,'' \emph{Frontiers in Human
  Neuroscience}, vol.~15, p. 799288, 2021.

\bibitem{zdunek2019linked}
R.~Zdunek, K.~Fona{\l}, and A.~Wo{\l}czowski, ``Linked cp tensor decomposition
  algorithms for shared and individual feature extraction,'' \emph{Signal
  Processing: Image Communication}, vol.~73, pp. 37--52, 2019.

\bibitem{kanatsoulis2018hyperspectral}
C.~I. Kanatsoulis, X.~Fu, N.~D. Sidiropoulos, and W.-K. Ma, ``Hyperspectral
  super-resolution: A coupled tensor factorization approach,'' \emph{IEEE
  Transactions on Signal Processing}, vol.~66, no.~24, pp. 6503--6517, 2018.

\bibitem{li2018fusing}
S.~Li, R.~Dian, L.~Fang, and J.~M. Bioucas-Dias, ``Fusing hyperspectral and
  multispectral images via coupled sparse tensor factorization,'' \emph{IEEE
  Transactions on Image Processing}, vol.~27, no.~8, pp. 4118--4130, 2018.

\bibitem{xu2024coupled}
T.~Xu, T.-Z. Huang, L.-J. Deng, J.-L. Xiao, C.~Broni-Bediako, J.~Xia, and
  N.~Yokoya, ``A coupled tensor double-factor method for hyperspectral and
  multispectral image fusion,'' \emph{IEEE Transactions on Geoscience and
  Remote Sensing}, 2024.

\bibitem{sorensen2013coupled}
M.~S{\o}rensen and L.~De~Lathauwer, ``Coupled tensor decompositions for
  applications in array signal processing,'' in \emph{2013 5th IEEE
  International Workshop on Computational Advances in Multi-Sensor Adaptive
  Processing (CAMSAP)}.\hskip 1em plus 0.5em minus 0.4em\relax IEEE, 2013, pp.
  228--231.

\bibitem{sorensen2016multidimensional1}
------, ``Multidimensional harmonic retrieval via coupled canonical polyadic
  decomposition—part i: Model and identifiability,'' \emph{IEEE Transactions
  on Signal Processing}, vol.~65, no.~2, pp. 517--527, 2016.

\bibitem{sorensen2016multidimensional2}
------, ``Multidimensional harmonic retrieval via coupled canonical polyadic
  decomposition—part ii: Algorithm and multirate sampling,'' \emph{IEEE
  Transactions on Signal Processing}, vol.~65, no.~2, pp. 528--539, 2016.

\bibitem{ermics2015link}
B.~Ermi{\c{s}}, E.~Acar, and A.~T. Cemgil, ``Link prediction in heterogeneous
  data via generalized coupled tensor factorization,'' \emph{Data Mining and
  Knowledge Discovery}, vol.~29, no.~1, pp. 203--236, 2015.

\bibitem{acar2015data}
E.~Acar, R.~Bro, and A.~K. Smilde, ``Data fusion in metabolomics using coupled
  matrix and tensor factorizations,'' \emph{Proceedings of the IEEE}, vol. 103,
  no.~9, pp. 1602--1620, 2015.

\bibitem{zhou2012fast}
G.~Zhou, A.~Cichocki, and S.~Xie, ``Fast nonnegative matrix/tensor
  factorization based on low-rank approximation,'' \emph{IEEE Transactions on
  Signal Processing}, vol.~60, no.~6, pp. 2928--2940, 2012.

\bibitem{zhang2016fast}
Y.~Zhang, G.~Zhou, Q.~Zhao, A.~Cichocki, and X.~Wang, ``Fast nonnegative tensor
  factorization based on accelerated proximal gradient and low-rank
  approximation,'' \emph{Neurocomputing}, vol. 198, pp. 148--154, 2016.

\bibitem{parikh2014proximal}
N.~Parikh, S.~Boyd \emph{et~al.}, ``Proximal algorithms,'' \emph{Foundations
  and Trends{\textregistered} in Optimization}, vol.~1, no.~3, pp. 127--239,
  2014.

\bibitem{nesterov1983method}
Y.~E. Nesterov, ``A method for solving the convex programming problem with
  convergence rate o (1/k\^{} 2),'' in \emph{Dokl. akad. nauk Sssr}, vol. 269,
  1983, pp. 543--547.

\bibitem{beck2009fast}
A.~Beck and M.~Teboulle, ``A fast iterative shrinkage-thresholding algorithm
  for linear inverse problems,'' \emph{SIAM journal on imaging sciences},
  vol.~2, no.~1, pp. 183--202, 2009.

\bibitem{guan2012nenmf}
N.~Guan, D.~Tao, Z.~Luo, and B.~Yuan, ``Nenmf: An optimal gradient method for
  nonnegative matrix factorization,'' \emph{IEEE Transactions on Signal
  Processing}, vol.~60, no.~6, pp. 2882--2898, 2012.

\bibitem{xu2013block}
Y.~Xu and W.~Yin, ``A block coordinate descent method for regularized
  multiconvex optimization with applications to nonnegative tensor
  factorization and completion,'' \emph{SIAM Journal on imaging sciences},
  vol.~6, no.~3, pp. 1758--1789, 2013.

\bibitem{xu2015alternating}
Y.~Xu, ``Alternating proximal gradient method for sparse nonnegative tucker
  decomposition,'' \emph{Mathematical Programming Computation}, vol.~7, no.~1,
  pp. 39--70, 2015.

\bibitem{cong2014low}
F.~Cong, G.~Zhou, P.~Astikainen, Q.~Zhao, Q.~Wu, A.~K. Nandi, J.~K. Hietanen,
  T.~Ristaniemi, and A.~Cichocki, ``Low-rank approximation based non-negative
  multi-way array decomposition on event-related potentials,''
  \emph{International journal of neural systems}, vol.~24, no.~08, p. 1440005,
  2014.

\bibitem{schenker2020flexible}
C.~Schenker, J.~E. Cohen, and E.~Acar, ``A flexible optimization framework for
  regularized matrix-tensor factorizations with linear couplings,'' \emph{IEEE
  Journal of Selected Topics in Signal Processing}, vol.~15, no.~3, pp.
  506--521, 2020.

\bibitem{hu2022discovering}
G.~Hu, H.~Li, W.~Zhao, Y.~Hao, Z.~Bai, L.~D. Nickerson, and F.~Cong,
  ``Discovering hidden brain network responses to naturalistic stimuli via
  tensor component analysis of multi-subject fmri data,'' \emph{NeuroImage},
  vol. 255, p. 119193, 2022.

\bibitem{wang2019fast}
X.~Wang, T.~Ristaniemi, and F.~Cong, ``Fast implementation of double-coupled
  nonnegative canonical polyadic decomposition,'' in \emph{ICASSP 2019-2019
  IEEE International Conference on Acoustics, Speech and Signal Processing
  (ICASSP)}.\hskip 1em plus 0.5em minus 0.4em\relax IEEE, 2019, pp. 8588--8592.

\bibitem{cichocki2009fast}
A.~Cichocki and A.-H. Phan, ``Fast local algorithms for large scale nonnegative
  matrix and tensor factorizations,'' \emph{IEICE transactions on fundamentals
  of electronics, communications and computer sciences}, vol.~92, no.~3, pp.
  708--721, 2009.

\bibitem{lee1999learning}
D.~D. Lee and H.~S. Seung, ``Learning the parts of objects by non-negative
  matrix factorization,'' \emph{Nature}, vol. 401, no. 6755, p. 788, 1999.

\bibitem{lee2009group}
H.~Lee and S.~Choi, ``Group nonnegative matrix factorization for {EEG}
  classification,'' in \emph{Artificial Intelligence and Statistics}, 2009, pp.
  320--327.

\bibitem{boyd2011distributed}
S.~Boyd, N.~Parikh, E.~Chu, B.~Peleato, J.~Eckstein \emph{et~al.},
  ``Distributed optimization and statistical learning via the alternating
  direction method of multipliers,'' \emph{Foundations and
  Trends{\textregistered} in Machine learning}, vol.~3, no.~1, pp. 1--122,
  2011.

\bibitem{georghiades2001few}
A.~S. Georghiades, P.~N. Belhumeur, and D.~J. Kriegman, ``From few to many:
  Illumination cone models for face recognition under variable lighting and
  pose,'' \emph{IEEE transactions on pattern analysis and machine
  intelligence}, vol.~23, no.~6, pp. 643--660, 2001.

\bibitem{cai2007spectral}
D.~Cai, X.~He, and J.~Han, ``Spectral regression for efficient regularized
  subspace learning,'' in \emph{2007 IEEE 11th international conference on
  computer vision}.\hskip 1em plus 0.5em minus 0.4em\relax IEEE, 2007, pp.
  1--8.

\bibitem{cong2012benefits}
F.~Cong, A.~H. Phan, Q.~Zhao, T.~Huttunen-Scott, J.~Kaartinen, T.~Ristaniemi,
  H.~Lyytinen, and A.~Cichocki, ``Benefits of multi-domain feature of mismatch
  negativity extracted by non-negative tensor factorization from eeg collected
  by low-density array,'' \emph{International journal of neural systems},
  vol.~22, no.~06, p. 1250025, 2012.

\bibitem{cong2013linking}
F.~Cong, V.~Alluri, A.~K. Nandi, P.~Toiviainen, R.~Fa, B.~Abu-Jamous, L.~Gong,
  B.~G. Craenen, H.~Poikonen, M.~Huotilainen \emph{et~al.}, ``Linking brain
  responses to naturalistic music through analysis of ongoing eeg and stimulus
  features,'' \emph{IEEE Transactions on Multimedia}, vol.~15, no.~5, pp.
  1060--1069, 2013.

\end{thebibliography}

\end{document}